\documentclass[runningheads]{llncs}

\usepackage{eccv}

\usepackage{eccvabbrv}

\usepackage{graphicx}
\usepackage{booktabs}

\usepackage[accsupp]{axessibility}  

\usepackage{hyperref}

\usepackage{orcidlink}
\usepackage[linesnumbered, ruled, vlined]{algorithm2e}
\SetKwInput{KwRequire}{Require}
\usepackage{verbatim}
\usepackage{xcolor}
\usepackage{wrapfig}
\definecolor{marking}{gray}{0.9}

\usepackage{eccvabbrv}

\usepackage{orcidlink}
\usepackage[accsupp]{axessibility}  

\usepackage[dvipsnames]{xcolor}
\usepackage{graphicx}
\usepackage{amsmath}
\usepackage{amssymb}
\usepackage{tabularx,booktabs}
\usepackage{color}
\usepackage{colortbl}
\usepackage{multirow}
\usepackage{multicol}
\usepackage{subcaption}
\usepackage{tabularray}
\UseTblrLibrary{diagbox}
\usepackage{pifont}
\usepackage{lipsum}  
\usepackage{makecell}
\usepackage{subcaption}
\usepackage{babel}
\usepackage{microtype}
\newcommand{\red}[1]{{\color{red}#1}}

\newcommand{\figref}[1]{Fig. \ref{#1}}

\newcommand{\secref}[1]{Sec. \ref{#1}}

\newcommand{\cmark}{\ding{51}}%
\newcommand{\xmark}{\ding{55}}%
\usepackage{tabularray}

\definecolor{LinkWater}{rgb}{0.85,0.882,0.949}

\begin{document}

\title{Enhancing Source-Free Domain Adaptive Object Detection with Low-confidence Pseudo Label Distillation}

\titlerunning{Low-confidence Pseudo Label Distillation (LPLD)}

\author{Ilhoon Yoon\inst{1}\orcidlink{0009-0008-5863-6374} \and
Hyeongjun Kwon\inst{1}\orcidlink{0009-0005-2424-7555} \and
Jin Kim\inst{1}\orcidlink{0009-0005-2710-7451} \and
Junyoung Park\inst{1}\orcidlink{0009-0008-9365-7234} \and \\
Hyunsung Jang\inst{1,2}\orcidlink{0000-0002-5797-7264} \and
Kwanghoon Sohn\inst{1,3}\orcidlink{0000-0002-3715-0331}
} 

\authorrunning{I.~Yoon et al.}


\institute{Yonsei University, Seoul \and LIG Nex1 \and KIST \\
\email{\{ilhoon231,~kwonjunn01,~kimjin928,~jun\_yonsei,
hyunsung.jang,~khsohn\}@yonsei.ac.kr}}

\maketitle
\begin{abstract}
 Source-Free domain adaptive Object Detection (SFOD) is a promising strategy for deploying trained detectors to new, unlabeled domains without accessing source data, addressing significant concerns around data privacy and efficiency.
 Most SFOD methods leverage a Mean-Teacher (MT) self-training paradigm relying heavily on High-confidence Pseudo Labels (HPL).
 However, these HPL often overlook small instances that undergo significant appearance changes with domain shifts. Additionally, HPL ignore instances with low confidence due to the scarcity of training samples, resulting in biased adaptation toward familiar instances from the source domain.
 To address this limitation, we introduce the Low-confidence Pseudo Label Distillation (LPLD) loss within the Mean-Teacher based SFOD framework.
 This novel approach is designed to leverage the proposals from Region Proposal Network (RPN), which potentially encompasses hard-to-detect objects in unfamiliar domains.
 Initially, we extract HPL using a standard pseudo-labeling technique and mine a set of Low-confidence Pseudo Labels (LPL) from proposals generated by RPN, leaving those that do not overlap significantly with HPL.
 These LPL are further refined by leveraging class-relation information and reducing the effect of inherent noise for the LPLD loss calculation.
 Furthermore, we use feature distance to adaptively weight the LPLD loss to focus on LPL containing a larger foreground area.
 Our method outperforms previous SFOD methods on four cross-domain object detection benchmarks.
 Extensive experiments demonstrate that our LPLD loss leads to effective adaptation by reducing false negatives and facilitating the use of domain-invariant knowledge from the source model.
 Code is available at \url{https://github.com/junia3/LPLD}.
  \keywords{
  Source-Free domain adaptive Object Detection}
\end{abstract}
\section{Introduction}
\label{sec:intro}
Object detection is a crucial task for advancing real-world applications like autonomous driving and robotics. 
Its success~\cite{fasterrcnn, yolo, detr, fpn, ssd} relies on annotated datasets with precise bounding boxes and category labels. 
However, trained detectors often suffer a significant performance drop in real-world scenarios due to the domain gap between training data (source domain) and deployment environments (target domain). 
Researchers have been addressing this with Unsupervised Domain Adaptive Object Detection (UDAOD), which reduces the domain gap by aligning the feature distributions of labeled source data and unlabeled target data~\cite{domain_wild, diversify, multi_adv, robust_udaod_mt, sw_od}, eliminating the need for labor-intensive annotation in the target domain.
\begin{figure}[t!]
  \centering
  \includegraphics[width=\textwidth]{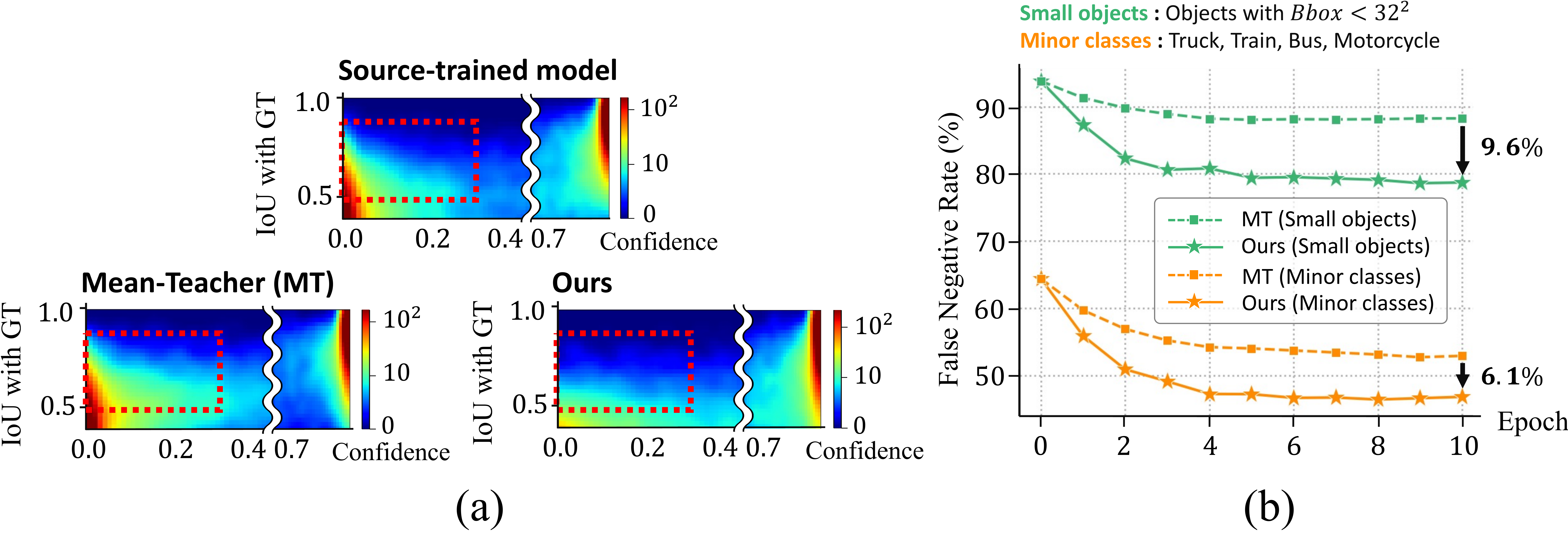}
  \caption{
  \textbf{(a) 2D Histogram of Proposals on Cityscapes}\cite{cityscapes}\textbf{ to Foggy Cityscapes}\cite{foggy}\textbf{.} Confidence Score and IoU with Ground Truth illustrates that before adaptation, the source-trained model often overlooks hard positive objects in the proposals with high IoUs but low confidence scores (\textbf{\textcolor{red}{red boxes}}).
  After adaptation, our LPLD loss promotes the detector to effectively capture hard positives with high-confidence scores in comparison to the Mean-Teacher (MT)~\cite{mean_teacher} based SFOD model which utilizes only the High-confidence Pseudo Labels (HPL).
  \textbf{(b) False Negative Rate (FNR) per Training Epoch.}
  Our model shows a consistently lower FNR than the MT baseline on \textit{hard-positive} objects (\eg, Minor classes, Small objects).
    }
    \label{fig:fig1}
\end{figure}

In practical applications, however, accessing source data is often restricted due to privacy, safety, and storage concerns~\cite{federated, membership, scaling}.
Even when accessible, transmitting large volumes of source data for every model deployment to new domains is more inefficient than sending only the source-trained model~\cite{model_adaptation, universalsfda, dowereallyneed}.
UDAOD methods are limited in these scenarios since they depend on simultaneous access to the labeled source data.
Given these constraints, Source-Free domain adaptive Object Detection (SFOD) emerges as a critical solution.
This limitation highlights the necessity for SFOD, adapting a source-trained model to an unlabeled target domain without any source data.
SFOD is more difficult than UDAOD, as it relies solely on the target domain images for adaptation.
Conventionally, SFOD methods~\cite{sfod_overlook, a2sfod, irg_sfda, periodically} follow a Mean-Teacher (MT)~\cite{mean_teacher} self-training paradigm.
The teacher detector generates pseudo labels with category scores and bounding boxes for weakly augmented target images, providing supervision for the student’s predictions on the same images with strong augmentation.
Then, the teacher detector parameters are updated as the Exponential Moving Average (EMA) of the updated student model parameters.
This MT framework enables adaptation without labels by leveraging teacher-student mutual learning.
On top of this MT framework, recent SFOD approaches have focused on enhancing target representation through styles~\cite{sfod_overlook}, dataset distributions~\cite{a2sfod}, and object relations~\cite{irg_sfda}, or improving adaptation stability~\cite{periodically}.

Despite their accuracy gain, they adopt the conventional pseudo-labeling process~\cite{pseudolabel, fixmatch} that depends on post-processing techniques like Non-Maximum Suppression (NMS) and score filtering, adopting a high-confidence threshold such as 0.9.
While this process ensures the reliability of pseudo-boxes, utilizing only the High-confidence Pseudo Labels (HPL) is problematic since the teacher detector often overlooks objects that the source-trained model struggles to detect.
For example, minor classes are often missing from HPL due to their low occurrences in the source dataset, making them vulnerable to domain variation. 
Similarly, small objects are frequently missing because they are inevitably hard to see and exhibit high visual variation with environmental changes such as fog.
With the label restriction of SFOD, this problem becomes more critical as the teacher detector consistently ignores the above \textit{hard-positive} objects during adaptation and the student detector takes biased supervision towards a few easily detectable objects.
This results in degraded overall performance.
To verify this issue, we visualize the histogram of confidence score and IoU with the Ground Truth of all proposals before post-processing in the target domain (Fig.~\ref{fig:fig1} (a)).
As highlighted in the red squares, the source-trained model has a large fraction of proposals containing \textit{hard positive} objects with low confidence scores.
These foreground proposals remain overlooked by typical MT methods (utilizing only HPL) even after the adaptation.
This oversight results in a high rate of false negatives during adaptation as depicted in Fig.\ref{fig:fig1} (b).
\begin{figure}[t!]
    \centering
    \includegraphics[width=\textwidth]{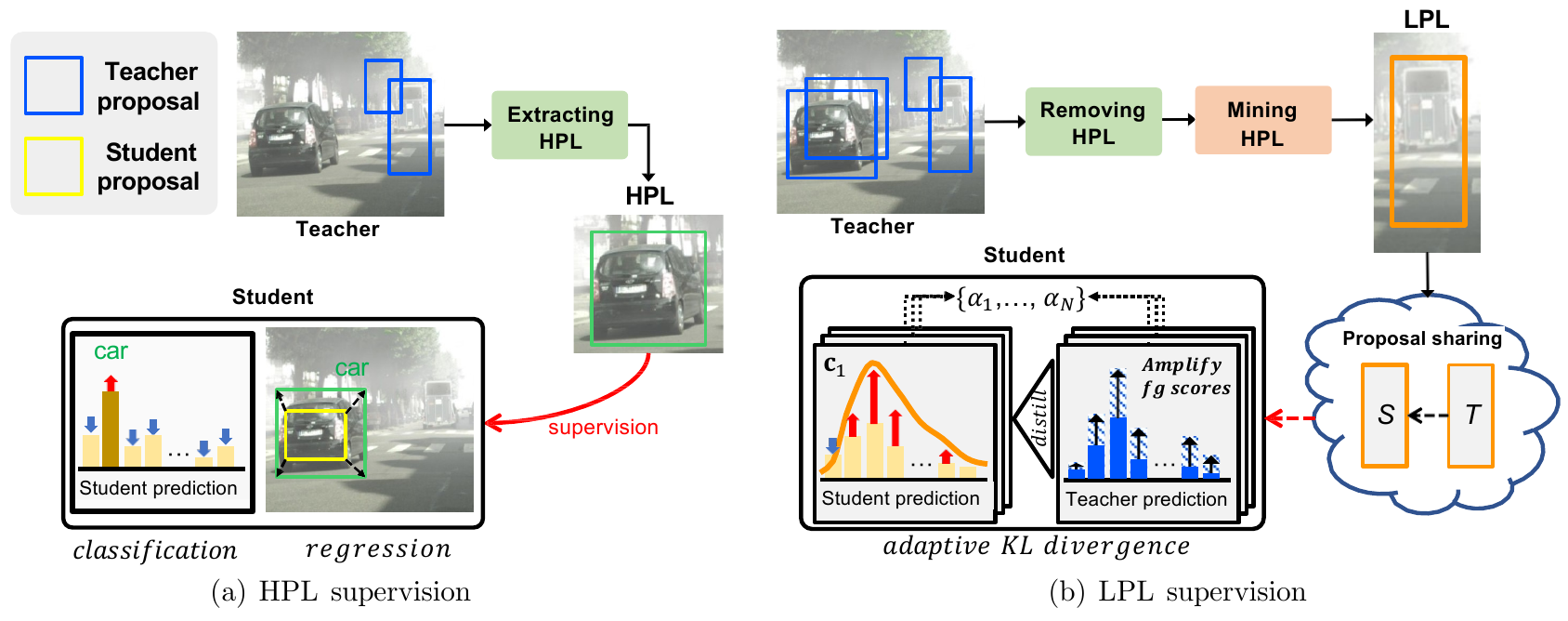}
    \caption{Comparison of two pseudo label supervision paradigms in Mean-Teacher SFOD methods. (a) High-confidence Pseudo Labels (HPL), which are provided as supervision for localization and classification to the student detector. (b) Low-confidence Pseudo Labels (LPL) supervise the student through our sophisticated distillation process.}
    \label{fig:fig2}
\end{figure}

To address this issue, we propose a Low-confidence Pseudo Label Distillation (LPLD) loss within the Mean-Teacher (MT) based SFOD framework.
This novel loss is designed to complement High Confidence Pseudo Labels (HPL) by capturing proposals that contain objects but have low confidence scores due to the domain gap.
Initially, we select HPL using a standard pseudo-labeling technique in Fig.~\ref{fig:fig2} (a).
We then exclude proposals highly overlapping with HPL before Non-Maximum Suppression (NMS), isolating a candidate pool of \textit{hard-positive} objects.
We refine these proposals by amplifying foreground category signals and applying the confidence threshold, producing Low-confidence Pseudo Labels (LPL).
Employing a Kullback-Leibler (KL) divergence loss between the teacher and student model predictions for each LPL helps mitigate inherent noise within LPL, as shown in Fig.~\ref{fig:fig2} (b).
Furthermore, we introduce adaptive weights to our LPLD loss based on teacher-student feature distance to focus on valuable LPL that are more likely to contain objects.
Our LPLD loss prevents the model from being biased towards \textit{easy-positive} objects in the target domain during adaptation, resulting in low false negatives as depicted in Fig.~\ref{fig:fig1} (b). 
It also achieves effective adaptation by promoting the usage of 
domain invariant knowledge from source-trained model with inter-class relations.
We summarize our contributions as follows:
\begin{itemize}
    \item We introduce a novel Low-confidence Pseudo Label Distillation (LPLD) loss on Mean-Teacher based Source-Free domain adaptive Object Detection (SFOD), through which we explore the frequently overlooked objects during conventional pseudo-labeling. 
    By leveraging the loss, the network gains a deeper understanding of the target domain by effectively utilizing false negatives and preventing bias toward \textit{easy-positive} objects.
    \item We introduce a feature-distance based adaptive weighting method for our LPLD loss to focus optimization on LPL that are more likely to contain objects and improve teacher-student consistency.
    \item The proposed method is evaluated on five domain-shift scenarios, comprising different types of domain-shift. Our method outperforms other SFOD counter
    parts on four domain shift scenarios and many UDAOD methods on all domain shift scenarios, demonstrating the effectiveness of our method.
\end{itemize}

\section{Related works}

\subsubsection{Unsupervised Domain Adaptation}

Unsupervised Domain Adaptation (UDA) utilizes labeled data from the source domain and unlabeled data from the target domain for domain adaptation in image classification.
UDA approaches can be broadly classified into three categories: feature alignment methods, image translation methods, and self-training methods.
Feature alignment methods~\cite{dann, uda_residual, uda_contrastive, uda_similarity, progressiveUDA, uda_maximum} aim to align the feature distributions between the source and target domain, making the features of the target domain similar to those of the source domain.
Image translation methods~\cite{cycada, i2i4da} translate source domain images into target domain-styled images utilizing given labels in the source-domain to transfer source domain knowledge to the target domain.
Self-training methods in UDA use pseudo labels generated on the target domain as supervision~\cite{uda_confidence, uda_generative, uda_selective, uda_cycle}.
In object detection, Unsupervised Domain Adaptation has been studied as Unsupervised Domain Adaptive Object Detection (UDAOD).
\cite{domain_wild, domain_harmonize} use instance-level and image-level representations to align feature distributions between the source and target domain. 
\cite{udaod_style_const, n2d_uda} apply an image translation approach to generalize knowledge and align the source and target domains. 
Furthermore, \cite{adaptive_st, ss_noisylabel} utilizes a self-training approach to generate instance-level pseudo labels for domain adaptation.
Although these methods have shown promising performance, all of them require access to the source domain data during target domain adaptation.

\subsubsection{Source-Free domain adaptive Object Detection}
In contrast to UDAOD, Source-Free domain adaptive Object Detection (SFOD) can only access the unlabeled target data and source-trained model dealing with the rising concerns about privacy and safety protection of the dataset\cite{federated, membership}.
To relieve the absence of source data, most SFOD works~\cite{freelunch, sfod_overlook, a2sfod, irg_sfda, periodically, exploiting_low_confidence} follow the self-training paradigm.
Li \etal~\cite{freelunch} is the pioneering work for SFOD, in which self-entropy descent is leveraged to obtain an appropriate confidence threshold for pseudo-labeling.
LODS~\cite{sfod_overlook} employs a style enhancement module and graph-based alignment to help the model learn domain-invariant features.
A$^2$SFOD~\cite{a2sfod} employs adversarial alignment on source-similar and dissimilar groups to facilitate representation capability. 
IRG~\cite{irg_sfda} integrates a graph convolutional network~\cite{gcn} to utilize object relations in contrast loss to enhance target representations. 
PETS~\cite{periodically} proposed periodic weight exchange between the static teacher and the student to mitigate error accumulation and enhance training stability.
More recently, LPU~\cite{exploiting_low_confidence} leverages proposals with confidence between low and high confidence threshold as soft pseudo labels utilizing contrastive loss to make the closest proposals' features similar.
Even though LPU proposes a way to utilize low-confidence proposals, it is hard to filter out clear background areas within low-confidence proposals. It results in simultaneously utilizing noisy foreground signals as labels, constantly learning and magnifying adverse signals during adaptation.
In this work, we refine the proposals for leveraging false negative candidates as LPL and introduce novel distillation loss to facilitate the network's understanding of the target domain while suppressing the inherent noise of LPL.

\section{Preliminaries}
\subsubsection{Problem statement}
Let \(\mathcal{D}_S = \{x_i, \mathcal{Y}_i\}^{N_S}_{i=1}\) represent the labeled source domain dataset, where \(x_i\) denotes the \(i^{th}\) image, and \(\mathcal{Y}_i\) is its corresponding label set containing object locations and class assignments, and \(\mathcal{D}_T = \{x_i\}^{N_T}_{i=1}\) denotes the target domain images.
\(N_{S}\) and \(N_{T}\) denote the number of the source and target domain images, respectively.
When deploying a model with source pre-trained parameters \(\mathrm{\Theta}_{pre}\) to an unseen domain, it is often challenging to access not only the target domain label but also the source dataset.
Thus, the goal of Source-Free domain adaptive Object Detection (SFOD) is to adapt the source model to the target domain without the aid of any source data \(\mathcal{D}_S\), utilizing only the pre-trained source model parameters \(\mathrm{\Theta}_{pre}\) and unlabeled target data \(\mathcal{D}_T\).

\subsubsection{Mean-Teacher based self-training framework}
Most of the recent advanced SFOD methods follow the Mean-Teacher (MT) self-training paradigm.
Generally, the teacher detector produces the pseudo label \(\hat{\mathcal{Y}}_{i}\) with weakly augmented image of \(x_i\) for supervising the student detector's predictions on strongly augmented image of \(x_i\).
The teacher detector is then updated by the optimized student detector's parameters via Exponential Moving Average (EMA).
Specifically, the pseudo label set \(\hat{\mathcal{Y}}_{i}\) is derived from the teacher detector's proposals \(\mathcal{P}_{i} = \{p_{i,j}\}^{N_{i}}_{j=1}\) through post-processing steps, including score filtering, Non-Maximum Suppression (NMS), and confidence thresholding.
\(N_{i}\) denotes the number of proposals according to the \(i^{th}\) target domain image \(x_{i}\).
Formally, the above MT-based framework is updated as follows:
\begin{equation}
    \begin{aligned}
    \mathcal{L}_{MT} = \mathcal{L}_{rpn}(x_{i}, &\hat{\mathcal{Y}}_{i}) + \mathcal{L}_{roi}(x_{i}, \hat{\mathcal{Y}}_{i}), \\
    \mathrm{\Theta}_{s} \leftarrow \mathrm{\Theta}_{s} - \eta\frac{\partial(\mathcal{L}_{MT})}{\partial \mathrm{\Theta}_{s}},& \quad \mathrm{\Theta}_{t} \leftarrow m\mathrm{\Theta}_{t} + (1-m)\mathrm{\Theta}_{s},
    \end{aligned}
    \label{eq:eq1}
\end{equation}
where \(\eta\), \(m\) denote the learning rate and teacher EMA rate, respectively. 
We denote the parameters of the teacher as \(\mathrm{\Theta}_{t}\) and those of the student detector as \(\mathrm{\Theta}_{s}\).
Note that, previous SFOD methods only adopt a high confidence threshold for pseudo labels such as 0.9, meaning that they only rely on high-confidence pseudo labels.

\begin{figure}[t!]
    \centering
    \includegraphics[width=\textwidth]{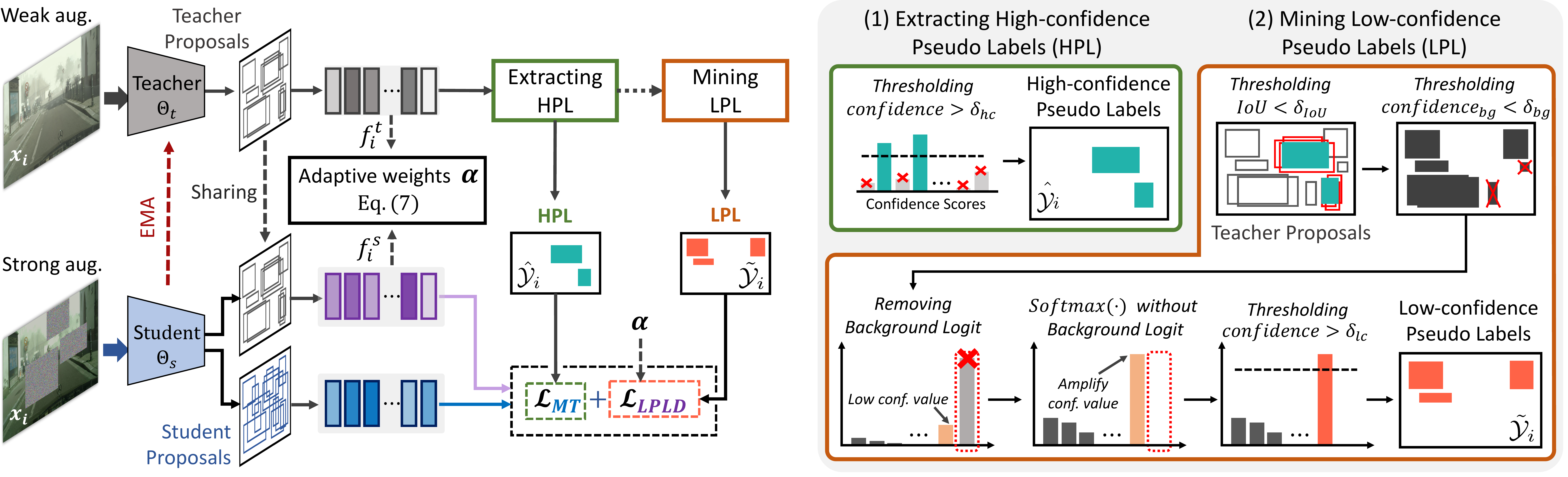}
    \caption{The overview of the proposed adaptive LPL Distillation framework.}
    \label{fig:frame}
\end{figure}

\section{Proposed method}
\subsection{Motivation and Overview}
While High-confidence Pseudo Labels (HPL) serve as reliable supervision for adaptation based on a high-confidence threshold, they are biased to overly confident instances.
We argue that solely leveraging HPL in SFOD methods restricts their representation capability within \textit{easy positive} instances, limiting adaptation performance for the target domain.

To tackle this challenge, we propose Low-confidence Pseudo Label Distillation (LPLD) to identify \textit{hard positive} instances and effectively learn their target domain representations, complementing HPL.
In particular, we first extract HPL via conventional pseudo-labeling algorithms~\cite{pseudolabel, fixmatch}.
Next, we exclude largely overlapped proposals with HPL from entire proposals, generating Low-confidence Pseudo Labels (LPL) where \textit{hard positive} instances are likely to be retained.
With LPL, we filter out background score, amplifying the foreground signals of remaining classes and apply threshold on foreground confidence.
These refined LPL are utilized as supervision for KL divergence loss on student predictions within corresponding regions of LPL (\secref{sec:lpld_process}).
Lastly, we calculate proposal-level teacher-student feature distances over the LPL region, thereby dynamically weighting KL divergence loss to prioritize the LPL containing more foreground (\secref{sec:AW}).
The overall procedure is illustrated in~\figref{fig:frame}.
In the following, we will explain each process in detail.

\subsection{Low-confidence Pseudo Label Distillation}
In this section, we elaborate on how our LPLD works to address the biased learning problem of the HPL-based method.\label{sec:lpld_process}
It is composed of three main processes: 1) \textbf{Extracting High-confidence Pseudo Labels} to find \textit{easy positive} instances, 2) \textbf{Mining Low-confidence Pseudo Labels} to identify missed \textit{hard positive} instances, and 3) \textbf{Low-confidence Pseudo Label Distillation loss} to stably improve the network's understanding of \textit{hard-positive} instances.
\subsubsection{Extracting High-confidence Pseudo Labels}\label{sec:HPL}

For each target image \(x_i\), we first obtain the proposal set \(\mathcal{P}_{i}\) from the Region Proposal Network (RPN) of the teacher detector.
Then, we employ a standard filtering process to the proposal set, including background score removal, score filtering, Non-Maximum Suppression (NMS) to obtain \(\bar{\mathcal{P}}_{i}\).
Then, we can obtain the HPL set \(\hat{\mathcal{Y}}_{i}\) by thresholding on confidence score as follows:
\begin{equation}
    \hat{\mathcal{Y}}_{i} = \{(\bar{p}_{i,j},\bar{\mathbf{c}}_{i,j}) | \bar{p}_{i,j} \in \bar{\mathcal{P}}_{i}, \max(\bar{\mathbf{c}}_{i,j}) > \delta_{hc}\},
    \label{eq:eq2}
\end{equation}
where \(j\) is proposal index, \(\delta_{hc}\) is the threshold of HPL, and $\max(\bar{\mathbf{c}}_{i,j})$ is the maximum class score from the class score vector $\bar{\mathbf{c}}_{i,j}$ of the filtered proposal.
Note that we can get HPL with high precision due to the high value of \(\delta_{hc}\).
The bounding boxes and class predictions of HPL are used to supervise the student detector with the regression loss \(\mathcal{L}_{reg}\) and classification loss \(\mathcal{L}_{cls}\).

\subsubsection{Mining Low-confidence Pseudo Labels}\label{sec:LPL}
To complement the HPL set with low-confidence proposals, we construct Low-confidence Pseudo Labels (LPL) set to capture \textit{hard positive} instances.

We first select the proposals that do not significantly overlap with HPL by thresholding on IoU (\eg less than 0.4) between the overall proposals and the HPL set.
Then we can get the candidate set of LPL $\widetilde{\mathcal{P}}_{i}$ as follows:
\begin{equation}
    \widetilde{\mathcal{P}}_{i} = \{p_{i,j} | p_{i,j}\in \mathcal{P}_{i}, \ \hat{p}_{i,k} \in \hat{\mathcal{Y}}_i, \ IoU(p_{i,j}, \hat{p}_{i,k}) < \delta_{IoU} \},
    \label{eq:eq3}
\end{equation}
where \(\delta_{IoU}\) is the overlapping IoU threshold, and \(\hat{p}_{i,k}\) is the bounding box for \(k^{th}\) pseudo label in \(\hat{\mathcal{Y}_i}\).
For a given candidate set \(\widetilde{\mathcal{P}}_{i}\) for LPL, a background confidence threshold \(\delta_{bg}\) filters out boxes that are certain the detection is background:
\begin{equation}
    \widetilde{\mathcal{P}}_{i}^{refined} = \{ p_{i,j} | p_{i,j}\in \widetilde{\mathcal{P}}_{i}, \ c_{i,j}^{bg} < \delta_{bg}\},
    \label{eq:eq4}
\end{equation}
where \(c_{i,j}^{bg}\) denotes the background score of proposal \(p_{i,j}\).
Next, we refine the proposals by removing the background score and dividing the foreground scores by L1-norm \(||\cdot||_{1}\) for amplifying their values, which is denoted as 
\(
\mathbf{c}^{amp}_{i,j} = \mathbf{c}_{i,j}^{fg}/{|| \mathbf{c}_{i,j}^{fg}||_{1}}
\).
Lastly, the LPL $\widetilde{\mathcal{Y}}_{i}$ are derived through thresholding the maximum class confidence of $\mathbf{c}^{amp}_{i,j}$ as follows:
\begin{equation}
    \widetilde{\mathcal{Y}}_{i} = \{(p_{i,j}, \mathbf{c}^{amp}_{i,j}) | p_{i,j} \in \widetilde{\mathcal{P}}_{i}^{refined},\ \max(\mathbf{c}^{amp}_{i,j}) > \delta_{lc}\},
    \label{eq:eq5}
\end{equation}
where \(\delta_{lc}\) is the LPL confidence threshold.

\subsubsection{Low-confidence Pseudo Label Distillation loss}\label{sec:LPLD}
Compared to HPL, LPL tends to contain \textit{hard positive} instances rather than \textit{easy positives}, indicating that they have lower reliability in localization and class predictions.
Therefore, utilizing LPL as supervision for classification and regression in the same manner as HPL impairs the student network's detection capabilities.
To address this problem, we employ the KL divergence loss between the student categorical prediction \(\mathbf{c}_{i,j}^{s}\) and amplified class distribution \(\widetilde{\mathbf{c}}_{i,j} \in \widetilde{\mathcal{Y}}_{i}\) in the same region of LPL.
Through LPLD, we provide solid representations of \textit{hard positive} instances for the student network, thereby enhancing the representation capability over the entire target domain.
The proposed LPLD loss can be formulated by:
\begin{equation}
    \mathcal{L}_{LPLD} = \frac{1}{| \widetilde{\mathcal{Y}}_{i} |}\sum_{\widetilde{\mathbf{c}}_{i,j}\in \widetilde{\mathcal{Y}}_{i}}D_{KL}(\mathbf{c}_{i,j}^{s} || \widetilde{\mathbf{c}}_{i,j}).
    \label{eq:eq6}
\end{equation}
Where $|\widetilde{\mathcal{Y}}_{i}|$ is the number of $\widetilde{\mathcal{Y}}_{i}$.
Note that, by optimizing the student model with our \(\mathcal{L}_{LPLD}\) loss, we can leverage inter-class relation between teacher and student detectors on the same region, leading to effectively utilizing the LPL set while avoiding the effect of inherent noise in LPL and preventing the bias towards \textit{easy-positive} objects.

\subsection{Adaptive weights for Distillation}\label{sec:AW}
\begin{wrapfigure}{r}{0.43\textwidth}
  \centering
  \vspace{-8mm}
  \includegraphics[width=0.42\textwidth]{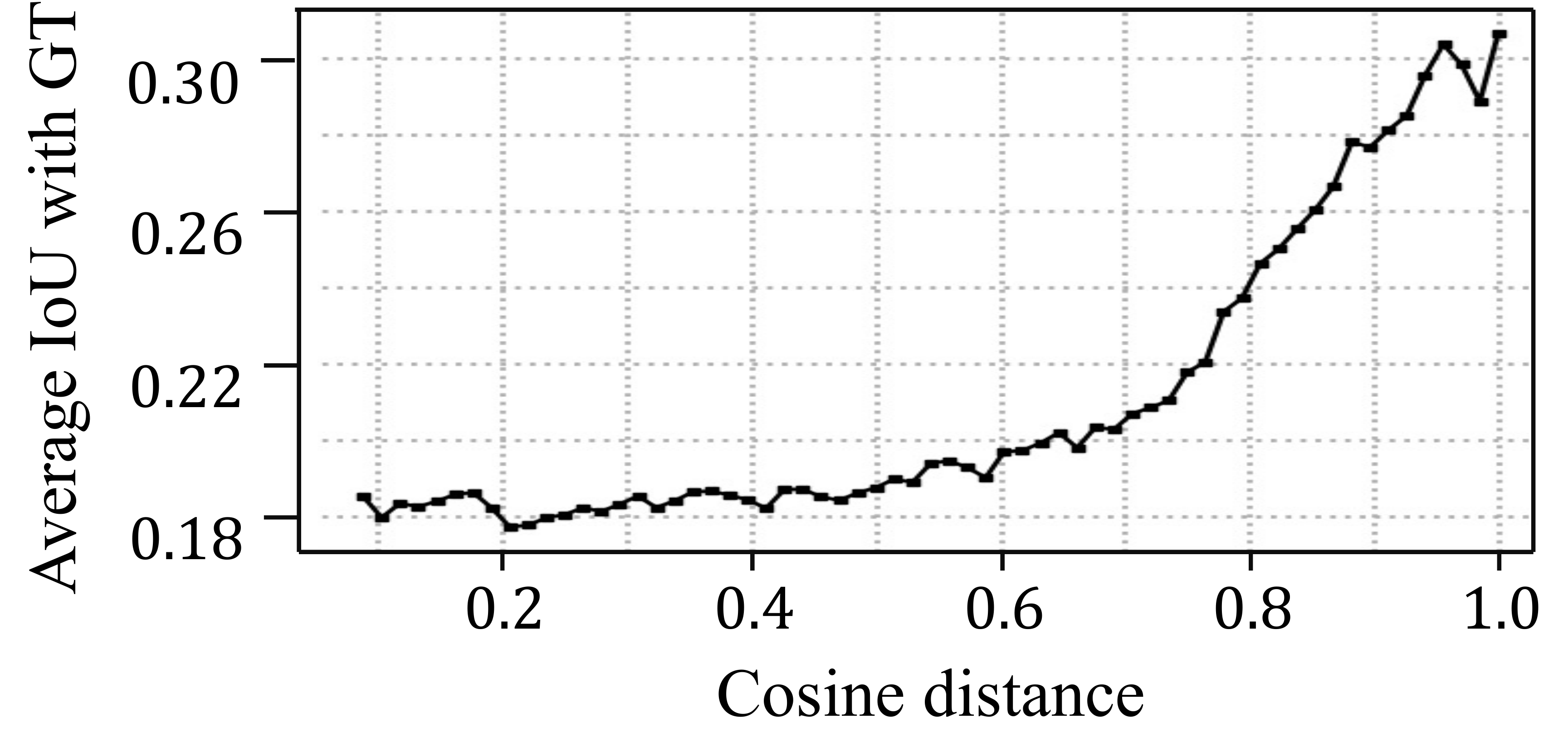}
  \vspace{-2mm}
  \caption{Average IoU with GT per feature distance.}
  \label{fig:fig4}
  \vspace{-12mm}
\end{wrapfigure}

We observe a positive correlation between the feature distance of the student and the teacher in the same region and the IoU with the ground-truth, as depicted in~\figref{fig:fig4}.
Therefore, we further improve LPLD loss by leveraging the feature distance.
Specifically, we utilize cosine distance between student and teacher feature for each LPL as adaptive weights \(\alpha\) to the KL divergence loss, enabling the network to prioritize learning on more object-dominated boxes rather than background and this can be formulated as:
\begin{equation}
    \alpha_{j} = 
    \begin{cases}
    d_{cos}(f^{s}_{i,j}, f^{t}_{i,j}), & \mbox{if }p_{i,j}\in \widetilde{\mathcal{Y}}_i, \\
    0, & otherwise.
    \end{cases}
    \label{eq:eq7}
\end{equation}
Where \(f^{t}_{i,j}\) and \(f^{s}_{i,j}\) represent the teacher's and student's features, extracted via RoI-Align process for the proposal \(p_{i,j}\). 
Finally, we apply the derived adaptive weights to its corresponding KL divergence loss terms and Eq.~\ref{eq:eq6} can be formulated as:
\begin{equation}
    \mathcal{L}_{LPLD} = \frac{1}{| \widetilde{\mathcal{Y}}_{i} |}\sum_{\widetilde{\mathbf{c}}_{i,j}\in \widetilde{\mathcal{Y}}_{i}}\alpha_{j}*D_{KL}(\mathbf{c}_{i,j}^{s} || \widetilde{\mathbf{c}}_{i,j}).
    \label{eq:eq8}
\end{equation}
Additionally, adaptive weights not only facilitate the optimization by focusing on the bounding boxes that are largely filled with the foreground objects but also enable the network to learn the robust representation by enhancing the consistency in teacher-student feature representations with the separate weak-strong augmentations.
\subsection{Total objectives}
Through the above procedures, we can formulate the total objectives as follows:
\begin{equation}
    \mathcal{L}_{total} = \mathcal{L}_{MT} + \mathcal{L}_{LPLD}.
    \label{eq:eq9}
\end{equation}
To sum up, \(\mathcal{L}_{MT}\) leverages HPL to improve the detection capability of the network with confidential prediction of \textit{easy positives}, whereas \(\mathcal{L}_{LPLD}\) on LPL makes the network focus on \textit{hard positive} instances, thereby 
providing solid understanding of target domain by focusing on \textit{hard positive} instances.

\section{Experiments}
To validate our method, we compare our result with state-of-the-art UDAOD, SFOD methods on five different domain shift scenarios, where these domain shifts can be divided into four types of domain shifts.

\subsection{Datasets}
A total of 7 datasets are used in the above domain shift scenarios, including the source domain dataset and the target domain dataset. 
1) \textbf{Cityscapes}\cite{cityscapes} is an urban street scene dataset that offers 5000 fine annotated images, where we use 2925 images as a training set and 500 images as a validation set.
2) \textbf{Foggy Cityscapes}\cite{foggy} is a dataset that has the synthetic fog applied to the Cityscapes dataset. Three versions of the Foggy Cityscapes exist by their visibility. 
3) \textbf{Sim10k}\cite{sim10k} is a synthetic dataset consisting of 10000 images of the street scene from the video game. It has computer-generated annotations of vehicles as alternatives to real-life data with manual annotation.
4) \textbf{KITTI}\cite{kitti} is a dataset consisting of 7481 training images. It is a street scene dataset similar to Cityscapes, but with different capturing environment, such as camera configuration, specification.
5) \textbf{Pascal-VOC}\cite{pascal} is a dataset consisting of real-world objects such as bird, cat, chair in various scenes.
6) \textbf{Clipart}\cite{clip_water} is an artistic dataset of clip arts consisting of 1000 images.
7) \textbf{Watercolor}\cite{clip_water} is an artistic dataset with watercolor paintings consisting of 2000 images.

\subsection{Implementation Details}
Following the SFOD setting from\cite{irg_sfda}, our baseline object detector is Faster R-CNN\cite{fasterrcnn} with a ResNet-50\cite{resnet} backbone pre-trained on ImageNet\cite{imagenet}, unless otherwise specified. Additionally, VGG-16\cite{vgg} is also used as the backbone network. For more details, please refer to supplementary materials.

\subsection{Quantitative Results}
On Tab.~\ref{tab:tab1}, \ref{tab:tab2}, \ref{tab:tab3} and \ref{tab:tab4}, we quantitatively compare our results with other methods in UDAOD, SFOD. Oracle is the baseline model trained with target data and its annotations, and source-only is the model trained on source domain data, evaluated on the target domain. In all evaluations, we use AP with IoU threshold 0.5 (AP50) as our evaluation metric.
\begin{table}[tb]
    \centering
    \caption{Quantitative mAP results for Cityscapes $\rightarrow$ Foggy Cityscapes. Minor classes are highlighted in \textbf{bold}.}
    \label{tab:tab1}
    \resizebox{0.7\textwidth}{!}{
    \begin{tabular}{cccccccccccc} 
     \toprule
     Type & Method & Backbone & Person & Rider &Car & \textbf{Truck} & \textbf{Bus} & \textbf{Train} & \textbf{Motor} & Bicycle & mAP \\
     \hline\hline
     \multirow{2}{*}{Source} & Source only & ResNet-50 & 29.3 & 34.1 & 35.8 & 15.4 & 26.0 & 9.1 & 22.4 & 29.7 & 25.2\\
      & Source only & VGG-16 & 29.7 & 36.7 & 36.5 & 13.9 & 30.7 & 5.0 & 20.1 & 32.7 & 25.7\\
     \hline
      \multirow{8}{*}{~~UDAOD~~} & DA Faster\cite{domain_wild} & ResNet-50 & 25.0 & 31.0 & 40.5 & 22.1 & 35.3 & 20.2 & 20.0 & 27.1 & 27.6\\
      & MAF\cite{multi_adv} & VGG-16 & 28.2 & 39.5 & 43.9 & 23.8 & 39.9 & 33.3 & 29.2 & 33.9 & 34.0\\
      & SWDA\cite{sw_od} & VGG-16 & 29.9 & 42.3 & 43.5 & 24.5 & 36.2 & 32.6 & 30.0 & 35.3 & 34.4\\
      & iFAN\cite{ifan} & VGG-16 & 32.6 & 48.5 & 22.8 & 40.0 & 33.0 & 45.5 & 31.7 & 27.9 & 35.3\\
      & CR DA\cite{categorical_od} & VGG-16 & 32.9 & 43.8 & 49.2 & 27.2 & 45.1 & 36.4 & 30.3 & 34.6 & 37.4\\
      & MeGA-CDA\cite{mega_cda} & VGG-16 & 37.7 & 49.0 & 52.4 & 25.4 & 49.2 & 46.9 & 34.5 & 39.0 & 41.8\\
      & Unbiased DA\cite{unbiasedMT} &  VGG-16 & 33.8 & 47.3 & 49.8 & 30.0 & 48.2 & 42.1 & 33.0 & 37.3 & 40.4\\
      & PT\cite{probabilistic_od} & VGG-16 & 40.2 & 48.8 & 59.7 & 30.7 & 51.8 & 30.6 & 35.4 & 44.5 & 42.7\\
      \hline
      \multirow{10}{*}{SFOD} & MT\cite{mean_teacher} & ResNet-50 & 37.4 & 43.0 & 45.0 & 27.2 & 37.2 & 25.1 & 28.2 & 38.2 & 34.3\\ 
      & SFOD\cite{freelunch} & VGG-16 & 32.6 & 40.4 & 44.0 & 21.7 & 34.3 & 11.8 & 25.3 & 34.5 & 30.6 \\
      & SFOD-Mosaic\cite{freelunch} & VGG-16 & 33.2 & 40.7 & 44.5 & 25.5 & 39.0 & 22.2 & 28.4 & 34.1 & 33.5 \\
      & LODS\cite{sfod_overlook} & VGG-16 & 34.0 & 45.7 & 48.8 & 27.3 & 39.7 & 19.6 & 33.2 & 37.8 & 35.8\\
      & A$^2$SFOD\cite{a2sfod} & VGG-16 & 32.3 & 44.1 & 44.6 & 28.1 & 34.3 & 29.0 & 31.8 & 38.9 & 35.4 \\
      & IRG\cite{irg_sfda} & ResNet-50 & 37.4 & 45.2 & 51.9 & 24.4 & 39.6 & 25.2 & 31.5 & 41.6 & 37.1 \\
      & PETS\cite{periodically} (single level) & VGG-16 & \textbf{42.0} & 48.7 & 56.3 & 19.3 & 39.3 & 5.5 & 34.2 & 41.6 & 35.9 \\
      & LPU\cite{exploiting_low_confidence} & VGG-16 & 39.0 & \textbf{50.3} & 55.4 & 24.0 & 46.0 & 21.2 & 30.3 & \textbf{44.2} & 38.8 \\ 
     & \cellcolor{marking}\textbf{Ours} & \cellcolor{marking}ResNet-50 & \cellcolor{marking}38.3 & \cellcolor{marking}42.9 & \cellcolor{marking}52.5 & \cellcolor{marking}28.4 & \cellcolor{marking}42.1 & \cellcolor{marking}\textbf{43.9} & \cellcolor{marking}33.4 & \cellcolor{marking}41.8 & \cellcolor{marking}40.4 \\ 
     & \cellcolor{marking}\textbf{Ours} & \cellcolor{marking}VGG-16 & \cellcolor{marking}39.7 & \cellcolor{marking}49.1 & \cellcolor{marking}\textbf{56.6} & \cellcolor{marking}\textbf{29.6} & \cellcolor{marking}\textbf{46.3} & \cellcolor{marking}26.4 & \cellcolor{marking}\textbf{36.1} & \cellcolor{marking}43.6 & \cellcolor{marking}\textbf{40.9} \\ 
     \midrule
     \multirow{2}{*}{Target} & Oracle & ResNet-50 & 38.7 & 46.9 & 56.7 & 35.5 & 49.4 & 44.7 & 35.9 & 38.8 & 43.1\\
      & Oracle & VGG-16 & 41.6 & 53.5 & 60.5 & 30.3 & 52.4 & 26.6 & 37.8 & 44.1 & 43.4\\
     \bottomrule
    \end{tabular}
    }
\end{table}

\subsubsection{Cityscapes to Foggy Cityscapes}
In deploying object detection model to the real world applications like autonomous vehicles, it is crucial to recognize that domain shifts caused by adverse weather conditions can pose significant risks. 
Cityscapes to Foggy Cityscapes exemplifies a domain shift induced by the fog, and we use the foggy level 0.02, which has the least visibility among three versions. 
The result on Foggy Cityscapes after domain adaptation is shown in Tab.~\ref{tab:tab1}. 
Our method achieves the mAP of 40.4 with ResNet-50 backbone and 40.9 with VGG-16 backbone, outperforming other SFOD methods in this domain shift scenario.
Furthermore, our method achieved the highest mAP for minor classes (truck, bus, train, motorcycle) among SFOD methods, registering at 37.0 mAP, which is 6.6 mAP higher than the second-best model.
\subsubsection{Sim10k to Cityscapes}
Numerous efforts have been made to substitute human-annotated labels in real images with synthetic datasets and their automatically generated labels.
However, significant challenge arises due to the substantial domain difference between synthetic dataset and real-world dataset, making it difficult to deploy a model trained on synthetic dataset to the real-world.
Sim10k to Cityscapes addresses the domain shift between synthetic dataset and real-world dataset. Since Sim10k only has annotations for cars, we only use car category for Sim10k and Cityscapes. 
The result on Cityscapes is shown in Tab.~\ref{tab:tab2}. Our method shows AP of 49.4 on the car.

\subsubsection{KITTI to Cityscapes}
Nowadays, various datasets depict the same scene, such as the urban road. However, they vary significantly due to factors like where they are collected, camera specification, and setup.
This leads to a domain shift between datasets, where datasets capturing similar scenes differ substantially.
Both KITTI and Cityscapes focus on road scenes, but they showcase notable visual difference.
Experiment on adapting the model trained on KITTI to the Cityscapes is done only in the common category of car, which can be observed on Tab.~\ref{tab:tab2}.
Our method shows AP of 51.3 on the car, outperforming other SFOD methods on the task.

\begin{table}[t]
    \centering
    \caption{Quantitative AP of Car results for Sim10k $\rightarrow$ Cityscapes, KITTI $\rightarrow$ Cityscapes.}
    \label{tab:tab2}
        \resizebox{0.5\textwidth}{!}{
        \begin{tabular}{cccc} 
             \toprule
             \multirow{2}{*}{Type} & \multirow{2}{*}{Method} & (Sim10k $\rightarrow$ City) & (Kitti $\rightarrow$ City) \\ \cline {3-4}
             & & AP of Car & AP of Car \\
             \hline\hline
             Source & Source Only & 32.0 & 33.9\\
             \hline
              \multirow{11}{*}{~~UDAOD~~} & DA Faster\cite{domain_wild} & 38.9 & 38.5 \\
              & MAF\cite{multi_adv} & 41.1 & 41.0 \\
              & Robust DA\cite{robust_udaod_mt} & 42.5 & 42.9\\
              & SWDA\cite{sw_od} & 40.1 & 37.9\\
              & ATF\cite{atf} & 42.8 & 42.1\\
              & HTCN\cite{domain_harmonize} & 42.5 & - \\
              & Cycle DA\cite{cycleda} & 41.5 & 41.7\\
              & MeGA-CDA\cite{mega_cda} & 44.8 & 43.0\\
              & Unbiased DA\cite{unbiasedMT} & 43.1 & - \\
              & PT\cite{probabilistic_od} & 55.1 & 60.2 \\
             \hline
              \multirow{9}{*}{SFOD} & MT\cite{mean_teacher} & 39.7 & 41.2  \\
              & SFOD\cite{freelunch} &  42.3 & 43.6 \\
              & SFOD-Mosaic\cite{freelunch} & 42.9 & 44.6 \\
              & LODS\cite{sfod_overlook} & - &43.9\\
              & A$^2$SFOD\cite{a2sfod} & 44.0 & 44.9 \\
              & IRG\cite{irg_sfda} & 45.2 & 46.9 \\
              & PETS\cite{periodically} & \textbf{57.8} & 47.0 \\
              & LPU\cite{exploiting_low_confidence} & 47.3 & 48.4\\
              & \cellcolor{marking} \textbf{Ours} & \cellcolor{marking}49.4 & \cellcolor{marking}\textbf{51.3} \\ 
             \bottomrule
            \end{tabular}
            }
\end{table}
\begin{table}[tb]
    \centering
    \caption{Quantitative mAP results for Pascal-VOC $\rightarrow$ Clipart.}
    \label{tab:tab3}
    \resizebox{\textwidth}{!}{
    \begin{tabular}{ccccccccccccccccccccccc} 
     \toprule
     Type & Method & aero & bcycle & bird & boat & bottle & bus & car & cat & chair & cow & table & dog & horse & bike & prsn & plnt & sheep & sofa & train & tv & mAP\\ [0.5ex] 
     \hline\hline
     Source & Source only & 35.6 & 52.5 & 24.3 & 23.0 & 20.0 & 43.9 & 32.8 & 10.7 & 30.6 & 11.7 & 13.8 & 6.0 & 36.8 & 45.9 & 48.7 & 41.9 & 16.5 & 7.3 & 22.9 & 32.0 & 27.8 \\
     \hline
     \multirow{6}{*}{~~UDAOD~~} & DA Faster\cite{domain_wild} & 15.0 & 34.6 & 12.4 & 11.9 & 19.8 & 21.1 & 23.3 & 3.10 & 22.1 & 26.3 & 10.6 & 10.0 & 19.6 & 39.4 & 34.6 & 29.3 & 1.00 & 17.1 & 19.7 & 24.8 & 19.8\\
     & BDC Faster\cite{sw_od} & 20.2 & 46.4 & 20.4 & 19.3 & 18.7 & 41.3 & 26.5 & 6.40 & 33.2 & 11.7 & 26.0 & 1.7 & 36.6 & 41.5 & 37.7 & 44.5 & 10.6 & 20.4 & 33.3 & 15.5 & 25.6\\
     & ADDA\cite{adda} & 20.1 & 50.2 & 20.5 & 23.6 & 11.4 & 40.5 & 34.9 & 2.3 & 39.7 & 22.3 & 27.1 & 10.4 & 31.7 & 53.6 & 46.6 & 32.1 & 18.0 & 21.1 & 23.6 & 18.3 & 27.4\\
     & BSR\cite{self_training} & 26.3 & 56.8 & 21.9 & 20.0 & 24.7 & 55.3 & 42.9 & 11.4 & 40.5 & 30.5 & 25.7 & 17.3 & 23.2 & 66.9 & 50.9 & 35.2 & 11.0 & 33.2 & 47.1 & 38.7 & 34.0 \\
     & WST\cite{self_training} & 30.8 & 65.5 & 18.7 & 23.0 & 24.9 & 57.5 & 40.2 & 10.9 & 38.0 & 25.9 & 36.0 & 15.6 & 22.6 & 66.8 & 52.1 & 35.3 & 1.0 & 34.6 & 38.1 & 39.4 & 33.8 \\
     
     & CLDA\cite{curriculum} & 22.3 & 61.5 & 17.9 & 16.0 & 34.8 & 34.9 & 32.0 & 9.8 & 31.5 & 26.7 & 24.0 & 10.8 & 23.5 & 49.8 & 55.3 & 27.3 & 5.7 & 22.1 & 25.3 & 21.6 & 27.6 \\
     \hline
     \multirow{4}{*}{SFOD} &  MT\cite{mean_teacher} & \textbf{22.3} & 42.3 & 23.8 & 21.7 & 23.5 & 60.7 & 33.2 & 9.1 & 24.7 & 16.7 & 12.2 & 13.1 & 26.8 & \textbf{73.6} & 43.9 & 34.5 & 9.1 & 24.3 & 37.9 & 42.2 & 29.1  \\
     & PL\cite{mean_teacher} & 18.3 & 48.4 & 19.2 & 22.4 & 12.8 & 38.9 & 36.1 & 5.2 & \textbf{36.9} & \textbf{24.8} & \textbf{29.3} & 9.1 & \textbf{34.6} & 58.6 & 43.1 & 34.3 & 9.1 & 14.4 & 26.9 & 19.8 & 28.2 \\
     & SFOD\cite{freelunch} &  20.1 & 51.5 & 26.8 & \textbf{23.0} & 24.8 & \textbf{64.1} & 37.6 & \textbf{10.3} & 36.3 & 20.0 & 18.7 & 13.5 & 26.5 & 49.1 & 37.1 & 32.1 & 10.1 & 17.6 & \textbf{42.6} & 30.0 & 29.5 \\
     & IRG\cite{irg_sfda} & 20.3 & 47.3 & \textbf{27.3} & 19.7 & 30.5 & 54.2 & 36.2 & \textbf{10.3} & 35.1 & 20.6 & 20.2 & 12.3 & 28.7 & 53.1 & 47.5 & \textbf{42.4} & 9.1 & 21.1 & 42.3 & \textbf{50.3} & 31.5\\ 
      & \cellcolor{marking} \textbf{Ours} & \cellcolor{marking} 18.9 & \cellcolor{marking}\textbf{66.1}& \cellcolor{marking}25.6 & \cellcolor{marking}21.1 & \cellcolor{marking}\textbf{37.6} & \cellcolor{marking}61.7 & \cellcolor{marking}\textbf{45.4} & \cellcolor{marking}9.1 & \cellcolor{marking}33.7 & \cellcolor{marking}11.2 & \cellcolor{marking}20.5 & \cellcolor{marking}\textbf{14.5} & \cellcolor{marking}32.3 & \cellcolor{marking}55.6 & \cellcolor{marking}\textbf{57.0} & \cellcolor{marking}37.3 & \cellcolor{marking}\textbf{18.2} & \cellcolor{marking}\textbf{31.7} & \cellcolor{marking}39.5 & \cellcolor{marking}42.6 & \cellcolor{marking}\textbf{34.0}\\ 
     \bottomrule
    \end{tabular}
    }
\end{table}
\begin{table}[!t]
    \centering
    \caption{Quantitative mAP results for Pascal-VOC  $\rightarrow$ Watercolor. Minor classes
are highlighted in \textbf{bold}.}
    \label{tab:tab4}
    \resizebox{0.55\textwidth}{!}{
    \begin{tabular}{ccccccccc} 
         \toprule
         Type & Method & \textbf{Bike} & Bird & \textbf{Car} & \textbf{Cat} & \textbf{Dog} & Person & mAP \\ 
         \hline\hline
         Source & Source only & 68.8 & 46.8 & 37.2 & 32.7 & 21.3 & 60.7 & 44.6\\
         \hline
          \multirow{6}{*}{~~UDAOD~~} & DA Faster\cite{domain_wild} & 75.2 & 40.6 & 48.0 & 31.5 & 20.6 & 60.0 & 46.0\\
          & BDC Faster\cite{sw_od} & 68.6 & 48.3 & 47.2 & 26.5 & 21.7 & 60.5 & 45.5\\
          & ADDA\cite{adda} & 79.9 & 49.5 & 39.5 & 35.3 & 29.4 & 65.1 & 49.8 \\
          & BSR\cite{self_training} & 82.8 & 43.2 & 49.8 & 29.6 & 27.6 & 58.4 & 48.6\\
          & WST\cite{self_training} & 77.8 & 48.0 & 45.2 & 30.4 & 29.5 & 64.2 & 49.2\\
          & HTCN\cite{domain_harmonize} & 78.6 & 47.5 & 45.6 & 35.4 & 31.0 & 62.2 & 50.1\\
         \hline
          \multirow{5}{*}{SFOD} & MT\cite{mean_teacher} &  73.6 & 47.6 & 46.6 & 28.5 & 29.4 & 58.6 & 47.1 \\ 
          & PL\cite{mean_teacher} & 74.6 & 46.5 & 45.1 & 27.3 & 25.9 & 54.4 & 46.1 \\
          & SFOD\cite{freelunch} & 76.2 & 44.9 & 49.3 & 31.6 & 30.6 & 55.2 & 47.9 \\
          & IRG\cite{irg_sfda} & 75.9 & \textbf{52.5} & \textbf{50.8} & 30.8 & 38.7 & 69.2 & 53.0 \\
        & \cellcolor{marking}\textbf{Ours} & \cellcolor{marking}\textbf{86.0} & \cellcolor{marking}51.8 & \cellcolor{marking}49.6 & \cellcolor{marking}\textbf{32.9} & \cellcolor{marking}\textbf{40.0} & \cellcolor{marking}\textbf{70.8} & \cellcolor{marking}\textbf{55.2} \\ 
         \bottomrule
    \end{tabular}
    }
\end{table}

\subsubsection{Pascal-VOC to Clipart, Pascal-VOC to Watercolor} 
Pascal-VOC to Clipart and Pascal-VOC to Watercolor both assume a domain shift between realistic dataset to artistic dataset. They both exhibit significant domain gap in their overall style and the appearance of each objects. Tab.~\ref{tab:tab3} shows the result of Pascal-VOC to Clipart, which outperforms the SFOD counterparts with mAP of 34.0. Tab.~\ref{tab:tab4} shows the result of Pascal-VOC to Watercolor, also outperforming the SFOD counterparts with mAP of 55.2.

\begin{table}[t]
    \centering
    \caption{Ablation studies with each components 
    (\textbf{left}) and variations of LPLD loss function (\textbf{right}) on Foggy Cityscapes dataset.}
    \begin{subtable}[b]{0.54\linewidth}
        \centering
        \resizebox{\textwidth}{!}{
            \begin{tabular}{cl|c}
            \toprule
            \multicolumn{2}{c|}{Components} &~mAP~\\ \midrule
            \textbf{(\uppercase\expandafter{\romannumeral1})} & Source model & 25.2 \\
            \textbf{(\uppercase\expandafter{\romannumeral2})} & LPL & 30.9 \\
            \textbf{(\uppercase\expandafter{\romannumeral3})} & LPL+Adaptive weights & 31.7 \\
            \textbf{(\uppercase\expandafter{\romannumeral4})} & HPL (MT) & 34.3 \\
            \textbf{(\uppercase\expandafter{\romannumeral5})} & HPL+LPL & 38.9 \\
            \rowcolor{marking} \textbf{(\uppercase\expandafter{\romannumeral6})} &~\textbf{HPL+LPL+Adaptive weights (Ours)}~& \textbf{40.4}\\
            \bottomrule
            \end{tabular}
            }
    \end{subtable}
    \begin{subtable}[b]{0.23\linewidth}
        \centering
        \resizebox{\textwidth}{!}{
        \begin{tabular}{cc|c|c} 
             \toprule
             \multicolumn{2}{c|}{$\mathcal{L}_{LPLD}$} &Adaptive& \multirow{2}{*}{~mAP~} \\ \cline{1-2} 
             $\mathcal{L}_{cls}$ & $\mathcal{L}_{reg}$ &~weights ($\alpha$)~& \\ \midrule
             CE & \cmark & \xmark & 36.6 \\
             CE & \cmark & \cmark & 38.3 \\
             \midrule
             CE & \xmark & \xmark & 37.2 \\
             CE & \xmark & \cmark & 39.0 \\ \midrule
             KL & \xmark & \xmark & 38.9 \\
             \rowcolor{marking} KL & \xmark & \cmark & \textbf{40.4} \\
             \bottomrule
        \end{tabular}%
        }
    \end{subtable}
    \label{tab:tab5}
\end{table}
\subsection{Further Analysis}
\subsubsection{Ablation Studies}
We conduct ablation studies to analyze the effectiveness of each pseudo label and the adaptive weights based on feature similarity in LPL. 
Left of Tab.~\ref{tab:tab5} compares each pseudo label, where HPL shows better performance (+9.1 mAP) than using LPL only (+5.7 mAP), with using both pseudo labels giving further improvement (+13.7 mAP), demonstrating the importance of using both pseudo labels.
Utilizing adaptive weights on the LPL along with HPL showed the best performance (+15.2 mAP), proving the importance of adaptively utilizing LPL.

\subsubsection{Loss Function}
We compare various loss choices of LPLD loss by altering classification loss and regression loss on Tab.~\ref{tab:tab5}.
When utilizing cross-entropy as the classification loss, a performance decrease of 1.4 mAP is observed. 
This can be attributed to the fact that the proposals around inaccurately localized LPL may have very low IoU or no overlap with the foreground object.
Furthermore, adopting regression loss along with cross-entropy loss resulted in a further performance decrease of 2.1 mAP.
This shows that utilizing regression loss on the inaccurately localized LPL is a sub-optimal solution.
Employing LPL in the same manner as HPL without adaptive weights resulted in the most significant performance drop of 3.8 mAP. 
In all cases, using adaptive weights on the LPL consistently yield better performance.

\subsubsection{Hyperparameter Sensitivity}
We perform experiments on hyperparameters $\delta_{IoU}$, $\delta_{bg}$ and $\delta_{lc}$ from LPL mining.
As shown in Tab.~\ref{tab:tab6}, our method shows promising results on various hyperparameter settings.
By observing $\delta_{IoU}$, we can see that utilizing only the proposals with IoU below 0.4 with HPL has the highest mAP. This corresponds to motivation of our method on using proposals that are not assigned as pairs to the HPL when calculating $\mathcal{L}_{MT}$, which generally uses IoU of 0.5. $\delta_{bg}$ value of 0.99 shows that removing some of the proposals that are overconfident on background can be helpful. $\delta_{lc}$ value of 0.9 shows that using foreground class confident proposals, tend to be the most optimal choice. We conjecture that foreground class confident proposals are likely to contain information specific to that class, thus showing the best result.
For all other domain shift scenarios, we fix these hyperparameters with $\delta_{IoU}=0.4$, $\delta_{bg}=0.99$, $\delta_{lc}=0.9$.
\begin{table}[htb!]
    \centering
    \caption{Ablation studies with LPL mining hyperparameters.}
    \label{tab:tab6}
    \begin{subtable}[t]{0.32\linewidth}
        \centering
        \resizebox{0.95\textwidth}{!}{
            \begin{tabular}{c|cccccc} 
                \toprule
                \multicolumn{7}{c}{Overlapping IoU threshold $\delta_{IoU}$.} \\ \midrule
                 $\delta_{IoU}$ & $0.0$ & $0.1$ & $0.2$ & $0.3$ & \cellcolor{marking}$\textbf{0.4}$ & $0.5$ \\ \hline
                 mAP & 38.7 & 38.4 & 39.1 & 39.5 & \cellcolor{marking}\textbf{40.4} & 38.3 \\
                 \bottomrule
            \end{tabular}
            }
    \end{subtable}
    \begin{subtable}[t]{0.32\linewidth}
        \centering
        \resizebox{0.95\textwidth}{!}{
            \begin{tabular}{c|cccccc}
                \toprule
                 \multicolumn{7}{c}{Background confidence threhold $\delta_{bg}$.} \\ \midrule
                 $\delta_{bg}$ & $0.95$ & $0.96$ & $0.97$ & $0.98$ & \cellcolor{marking}$\textbf{0.99}$ & 1.00 \\ \hline
                 mAP & 38.3 & 38.6 & 38.8 & 39.4 & \cellcolor{marking} \textbf{40.4} & 39.3\\
                 \bottomrule
            \end{tabular}
        }
    \end{subtable}
    \begin{subtable}[t]{0.32\linewidth}
        \centering
        \resizebox{0.9\textwidth}{!}{
            \begin{tabular}{c|cccccc}
                \toprule
                 \multicolumn{7}{c}{LPL confidence threhold $\delta_{lc}$.} \\ \midrule
                 $\delta_{lc}$ & $0.4$ & $0.5$ & $0.6$ & $0.7$ & $0.8$ & \cellcolor{marking}$\textbf{0.9}$ \\ \hline
                 mAP & 38.1 & 38.6 & 38.7 & 38.8 & 39.0 & \cellcolor{marking}\textbf{40.4} \\ \bottomrule
            \end{tabular}
        }
    \end{subtable}
\end{table}

\begin{figure}[tb]
    \centering
\includegraphics[width=0.95\textwidth]{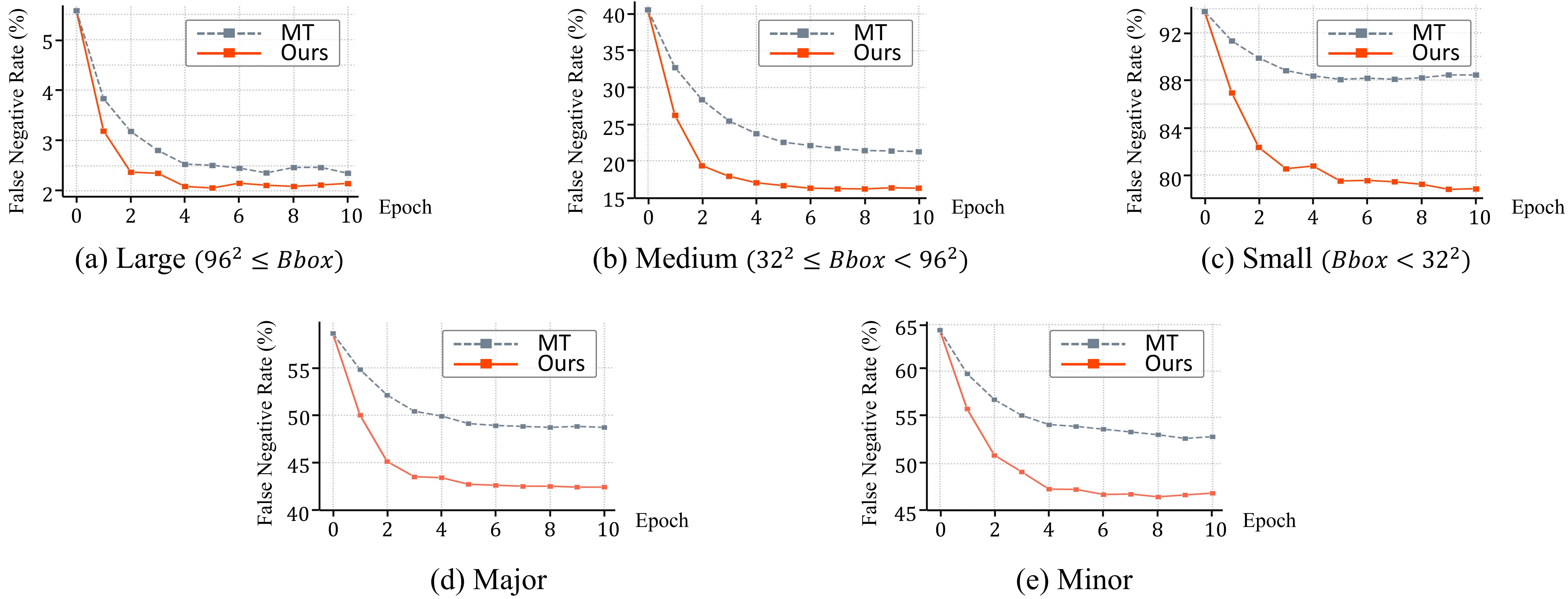}
    \caption{False Negative Rate (\%) per Training Epoch for (a) Large, (b) Medium, (c) Small object instances and (d) Major, (e) Minor class instances.}
    \label{fig:ablation}
\end{figure}
\begin{figure}[htb!]
    \begin{minipage}[h]{.245\linewidth}
    \centering
    \includegraphics[width=\linewidth]{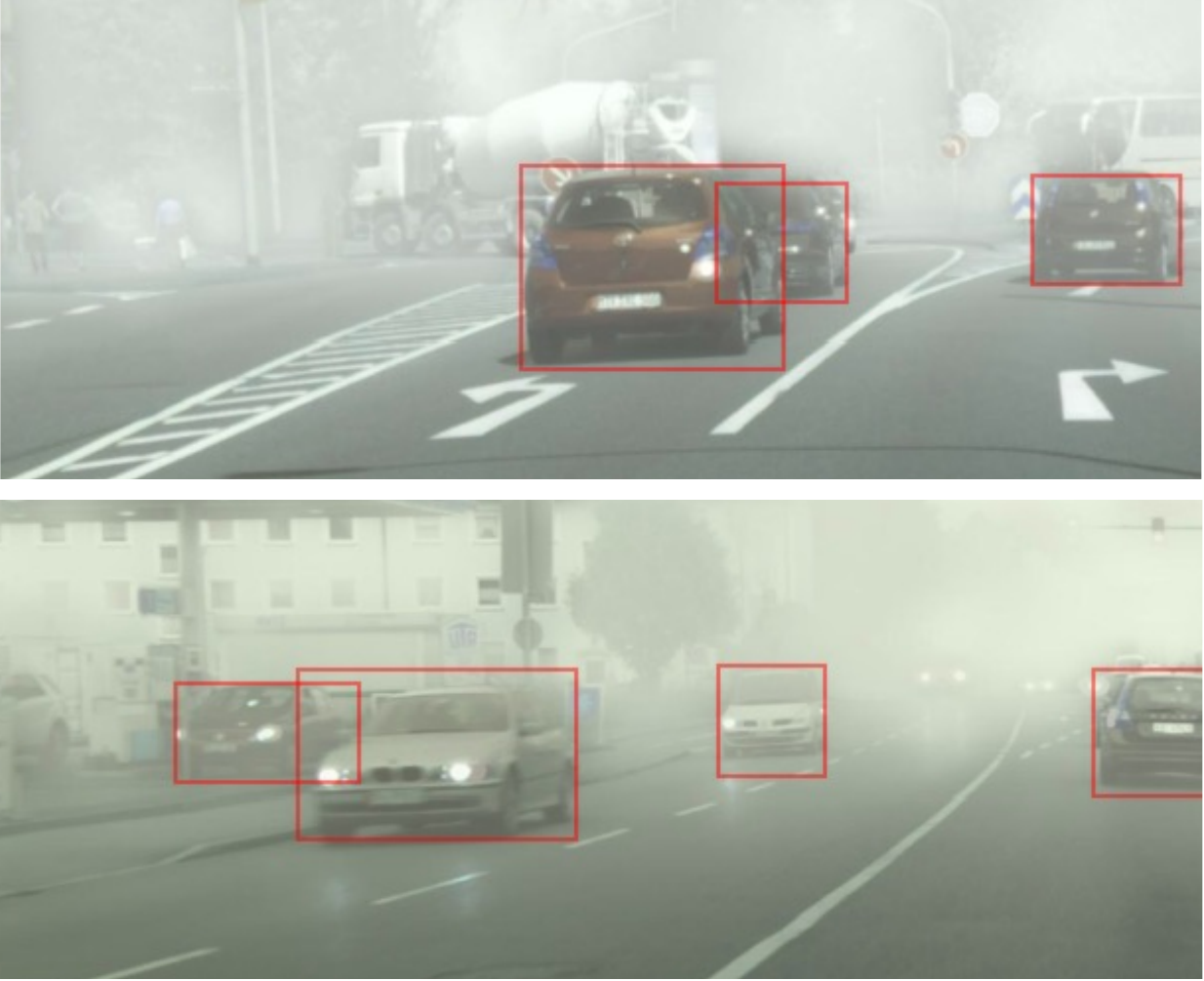}
    \caption*{\scriptsize{(a) Source}}
  \end{minipage}
  \begin{minipage}[h]{.245\linewidth}
    \centering
    \includegraphics[width=\linewidth]{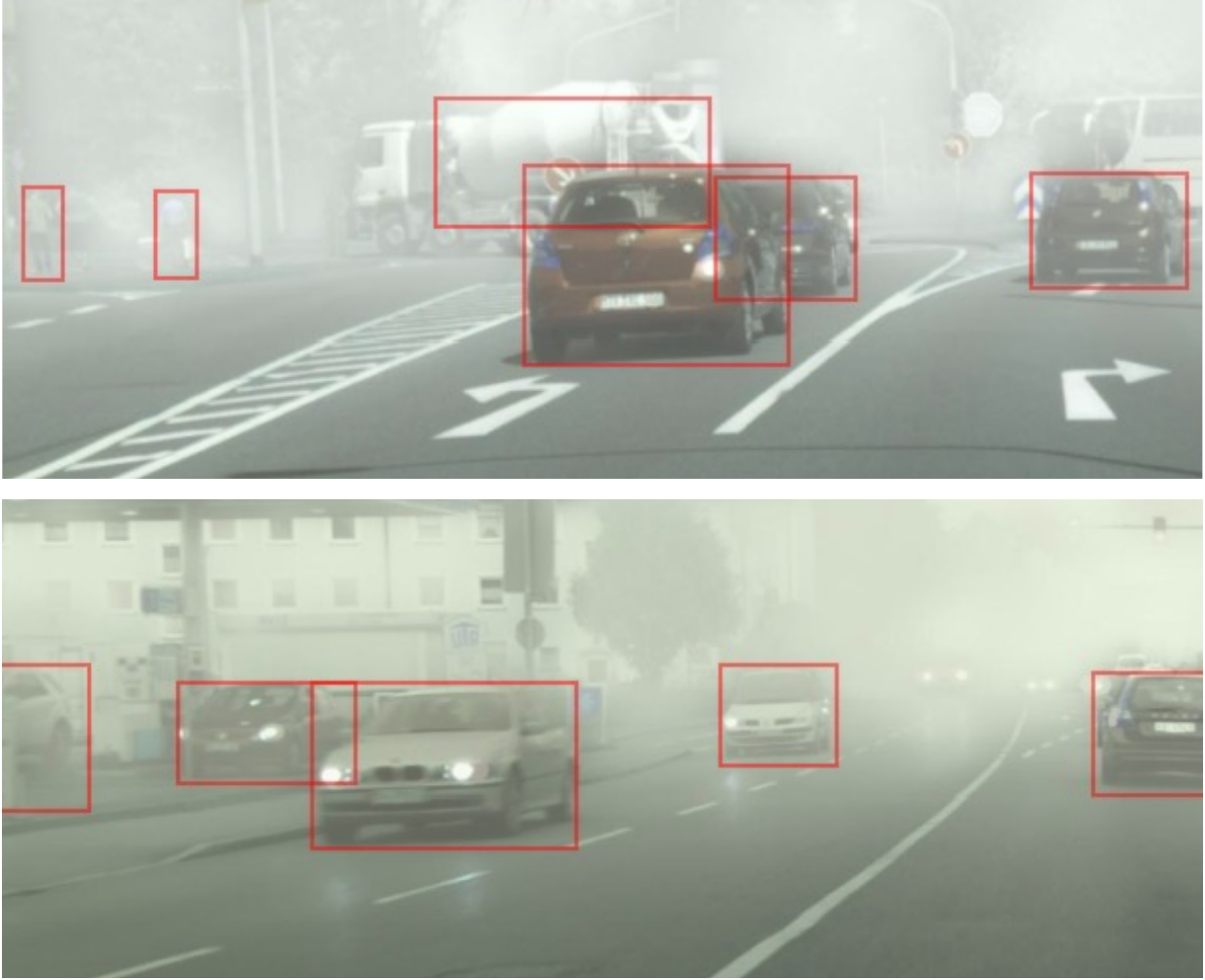}
    \caption*{\scriptsize{(b) MT(HPL)~\cite{mean_teacher}}}
  \end{minipage}
  \begin{minipage}[h]{.245\linewidth}
    \centering
    \includegraphics[width=\linewidth]{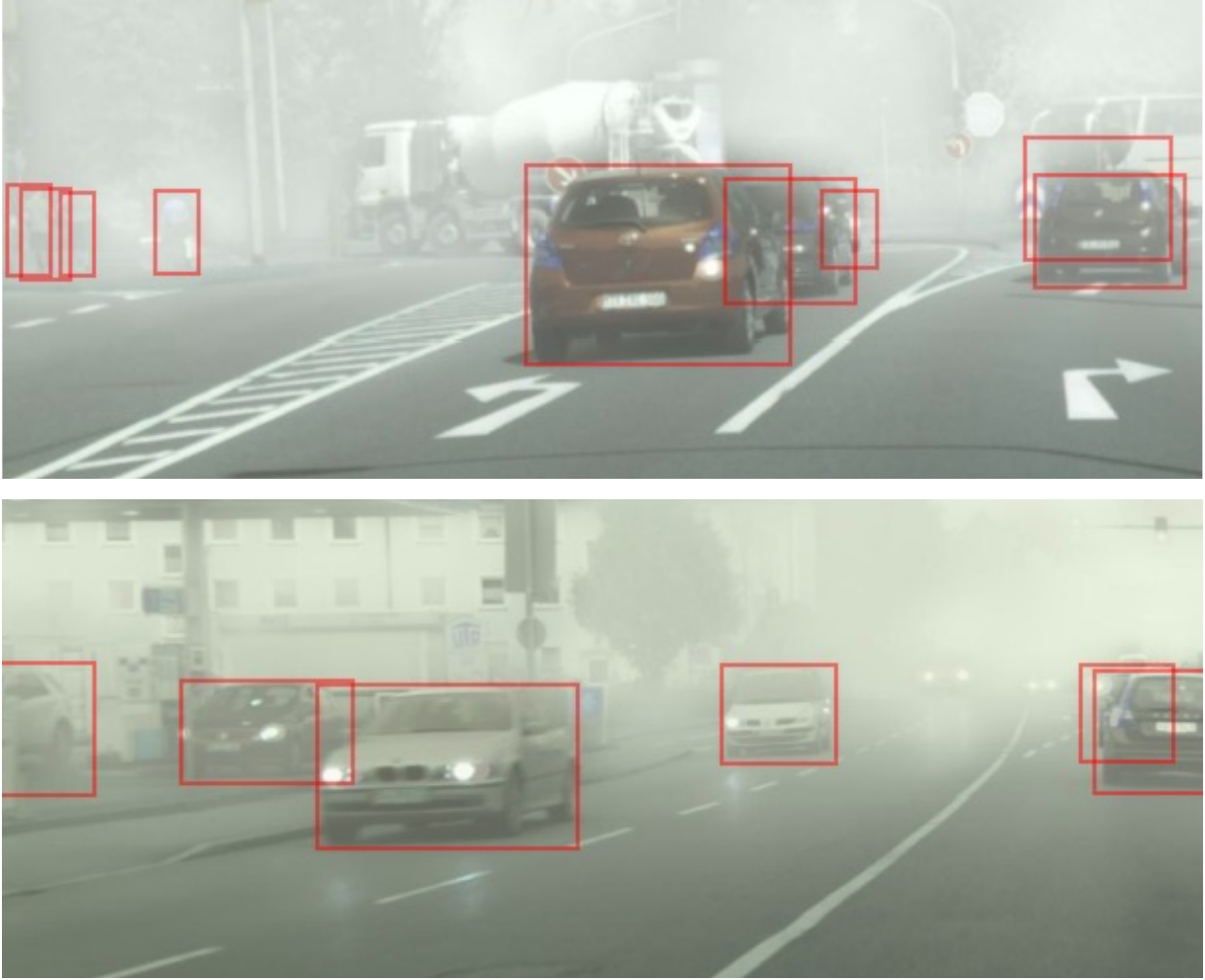}
    \caption*{\scriptsize{(c) IRG~\cite{irg_sfda}}}
  \end{minipage}
  \begin{minipage}[h]{.245\linewidth}
    \centering
    \includegraphics[width=\linewidth]{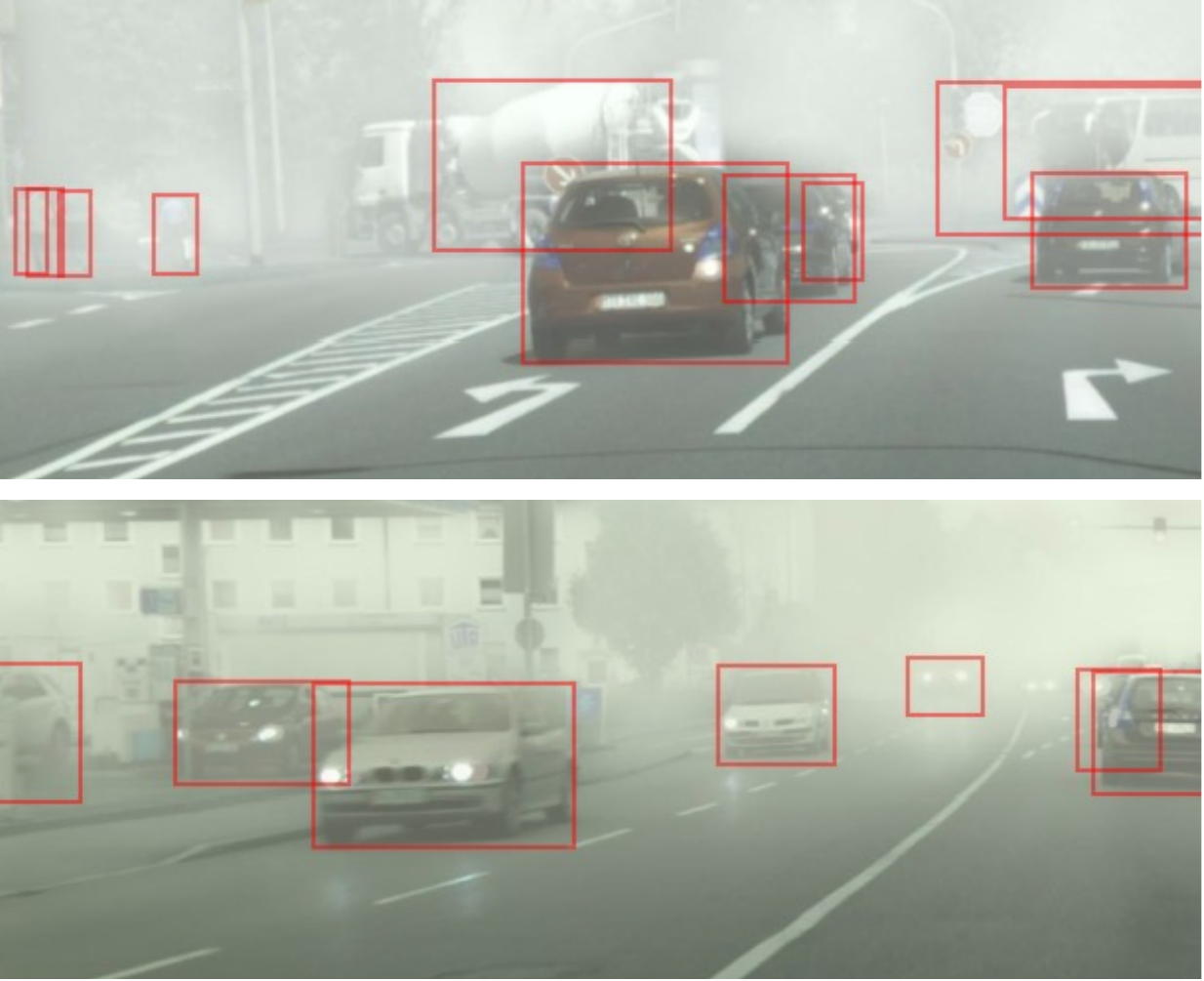}
    \caption*{\scriptsize{(d) Ours}}
  \end{minipage}
  \caption{Qualitative results for Cityscapes $\to$ Foggy Cityscapes. Bounding boxes in \textbf{\red{red}} refer to the prediction. \textit{Zoom in for best view.}}
  \label{fig:qual}
\end{figure}

\subsubsection{False Negative Rate Difference}
We conducted ablations on the False Negative Rate (FNR) using MT and our method across objects of varying sizes and classes~\figref{fig:ablation}. 
Our method consistently lowers FNR compared to the MT. 
Specifically, for large, medium, and small objects, our approach reduces FNR by 0.23\%, 5.02\%, and 9.58\%, respectively. 
For major (person, rider, car, bicycle) and minor (truck, bus, train, motorcycle) classes, our approach shows a reduction of 6.10\% and 6.31\%, respectively.

\subsubsection{Qualitative Analysis}
Additionally, we provide some qualitative results in~\figref{fig:qual}.
We compare our method with the source-trained model, HPL-based method, and IRG~\cite{irg_sfda}.
As can be seen, our method outperforms other methods in terms of prediction quality.
It is noteworthy that our method has improved the ability to detect \textit{hard positives}, including those from rare classes or small-size instances, such as trucks, buses, and highly occluded small cars in~\figref{fig:qual}.
Please refer to supplementary materials for qualitative analysis on various domain shift scenarios.

\section{Conclusions}
We introduce a novel strategy to enhance Source-Free domain adaptive Object Detection focusing on effectively utilizing Low-confidence Pseudo Labels (LPL), termed LPLD loss.
Our method effectively reduces the false negative rate in new domains, improving the model's adaptability and performance.
Experiments demonstrate that our approach improves adaptation accuracy across various domain shift scenarios, suggesting that even low-confidence proposals have valuable information that cannot be overlooked.

\section*{Acknowledgements}
This research was supported by the National Research Foundation of Korea (NRF) grant funded by the Korea government (MSIT) (NRF2021R1A2C2006703), and partly supported by the Yonsei Signature Research Cluster Program of 2024 (2024-22-0161).
%
\bibliographystyle{splncs04}
\bibliography{egbib}
\renewcommand\thesection{\Alph{section}}
\title{Enhancing Source-Free Domain Adaptive Object Detection with Low-confidence Pseudo Label Distillation\\ - Supplementary Materials - } 

\titlerunning{Low-confidence Pseudo Label Distillation (LPLD)} 

\author{} 

\authorrunning{I.~Yoon et al.}
\institute{\email{}}

\maketitle
In this document, we present our LPLD algorithm in Sec.~\ref{sec:alg} and more implementation details in Sec.~\ref{sec:imple}.
Also, we provide full 2D histogram visualizations and further analysis on the performance gain and the mining process of Low-confidence Pseudo Labels (LPL) in Sec. \ref{sec:suppl}.
Moreover, we present qualitative results on various cross-domain object detection benchmarks in Sec.~\ref{sec:supplequal}.
\section{Algorithm of LPLD}\label{sec:alg}
To facilitate understanding, we present the pseudo-code of our proposed method, as shown in Algorithm~\ref{al_LPLD}.
Note that all references in Algorithm~\ref{al_LPLD} are brought from the equations presented in the main paper.
\setlength{\textfloatsep}{-20pt}
\begin{algorithm}[h]
    \caption{\footnotesize{LPLD Algorithm}}
    \label{al_LPLD}
    \LinesNumbered
    
    \KwRequire{\footnotesize
    Student backbone $\mathcal{F}^s$, RPN $\mathcal{G}^s$ and detector $\mathcal{H}^s$, \\
    Teacher backbone $\mathcal{F}^t$, RPN $\mathcal{G}^t$ and detector $\mathcal{H}^t$,
    Unlabeled target dataset $\mathcal{D}_{T}$, Source pre-trained parameter $\mathrm{\Theta}_{pre}$, 
    Strong aug. $\mathcal{A}^{strong}(\cdot)$, Weak aug. $\mathcal{A}^{weak}(\cdot)$. \vspace{-15pt}
    \\\hrulefill
    }
    Initialize $\mathrm{\Theta}_{t} \leftarrow \mathrm{\Theta}_{pre}$, $\mathrm{\Theta}_{s} \leftarrow \mathrm{\Theta}_{pre}$. \\
    \For{each epoch}{
        \For{$0 \le i < N_T$}{
            Extract an \textit{\iid} sample $x_i$ from $\mathcal{D}_{T}$. \\
            ($x_i^{weak},~x_i^{strong}) \leftarrow (\mathcal{A}^{weak}(x_i),~\mathcal{A}^{strong}(x_i)$) \\
            \textrm{\textbf{With Teacher Model :}}\\
            \quad $f_i^t \leftarrow \mathcal{F}^t(x_i^{weak})$, $\mathcal{P}_i \leftarrow \mathcal{G}^t(f_i^t)$, $\bar{\mathcal{P}}_i \leftarrow \operatorname{NMS}(\mathcal{P}_i)$  \\
            \quad Yield HPL $\hat{\mathcal{Y}_i}$ according to Eq.~(\ref{eq:eq2}).\\
            \quad Find $\widetilde{\mathcal{P}}_i$ by applying IoU threshold according to Eq.~(\ref{eq:eq3}).\\
            \quad Find $\widetilde{\mathcal{P}}_i^{refined}$ by filtering background according to Eq.~(\ref{eq:eq4}).\\
            \quad Yield LPL $\widetilde{\mathcal{Y}}_i$ by utilizing Reweigted conf. according to Eq.~(\ref{eq:eq5}).\\

            \textrm{\textbf{With Student Model :}}\\
            \quad $f^s_i \leftarrow \mathcal{F}^s(x_i^{strong})$ \\
            \quad Compute $\mathcal{L}_{MT}$ according to Eq.~(\ref{eq:eq1}).
            \\
            \quad $\{f^{s}_{i,j}\}_{j=1}^{\vert \widetilde{\mathcal{Y}}_i \vert},~\{f^{t}_{i,j}\}_{j=1}^{\vert \widetilde{\mathcal{Y}}_i \vert} \leftarrow \operatorname{RoIAlign}(f^s_i, \widetilde{\mathcal{Y}}_i),~\operatorname{RoIAlign}(f^t_i, \widetilde{\mathcal{Y}}_i)$ \\
            \quad Compute feature distance $\alpha_j$ between $f^{s}_{i,j}, f^{t}_{i,j}$ according to Eq.~(\ref{eq:eq7}).\\
            \quad Compute $\mathcal{L}_{LPLD}$ according to Eq.~(\ref{eq:eq6}).\\
            \quad Apply $\alpha_j$ as an adaptive weight to $\mathcal{L}_{LPLD}$ according to Eq.~(\ref{eq:eq8}).\\
            \quad Update $\mathrm{\Theta}_{s}$ by gradient descent with $\mathcal{L}_{MT}$ and $\mathcal{L}_{LPLD}$\\
        }
        Update $\mathrm{\Theta}_{t}$ by EMA rate with $\mathrm{\Theta}_{s}$
    }
\end{algorithm}
\setlength{\textfloatsep}{0pt}
\section{Implementation details}\label{sec:imple}
Following the SFOD setting from\cite{irg_sfda}, our baseline object detector is Faster R-CNN\cite{fasterrcnn}, utilizing a ResNet-50\cite{resnet} backbone pre-trained on ImageNet\cite{imagenet}, unless stated otherwise. Additionally, we also employ VGG-16\cite{vgg} as the backbone network.
For optimizer, we use SGD with learning rate 0.001, momentum 0.9, weight decay 0.0001.
Resizing is done for all images to have the shorter edge of 600 and longest edge of maximum 1333, with keeping its ratio.
Weak augmentation consists of resizing, while strong augmentation includes additional techniques such as color jitter, grayscale conversion, Gaussian blur, and random erasing.
The batch size is set to 1.
EMA rate in the teacher is set to 0.75, and HPL are generated from the teacher and filtered with the confidence threshold of 0.7.
\section{Additional Analysis}\label{sec:suppl}
\subsection{Full visualization of 2D histogram}
\begin{figure}[htb!]
    \centering
    \includegraphics[width=\textwidth]{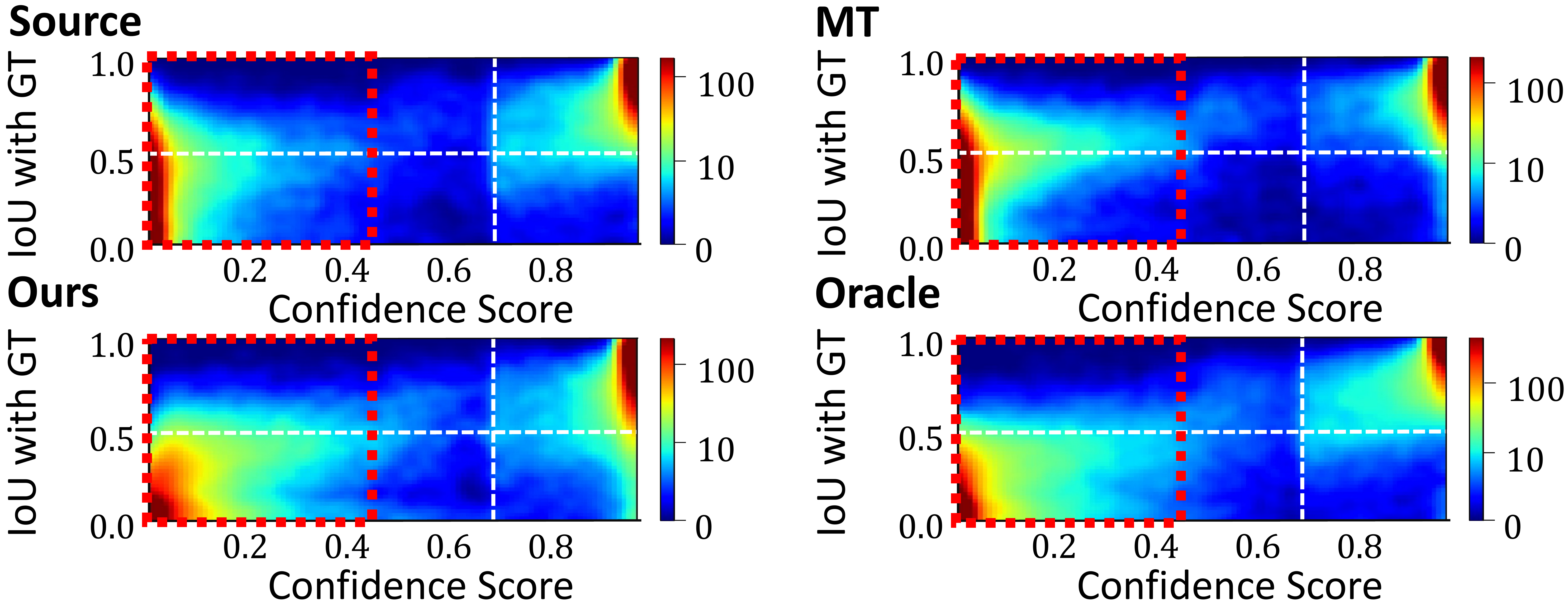}
    \caption{
    2D Histogram of Proposals on Cityscapes $\rightarrow$ Foggy Cityscapes.
    }
    \label{fig:full_2d}
\end{figure}

We fully visualize the 2D histogram of the proposals in Fig.~\ref{fig:full_2d}, showing all proposals and the target-only trained model (Oracle). 
Unlike the source and the Mean-Teacher (MT) baselines, our model achieves high confidence in proposals with an IoU $>$ 0.5 and low confidence for the other side, similar to the Oracle model.

\subsection{Performance gain and number of instances for each class on Cityscapes to Foggy Cityscapes}
\begin{figure}[htb!]
    \centering
    \includegraphics[width=\textwidth]{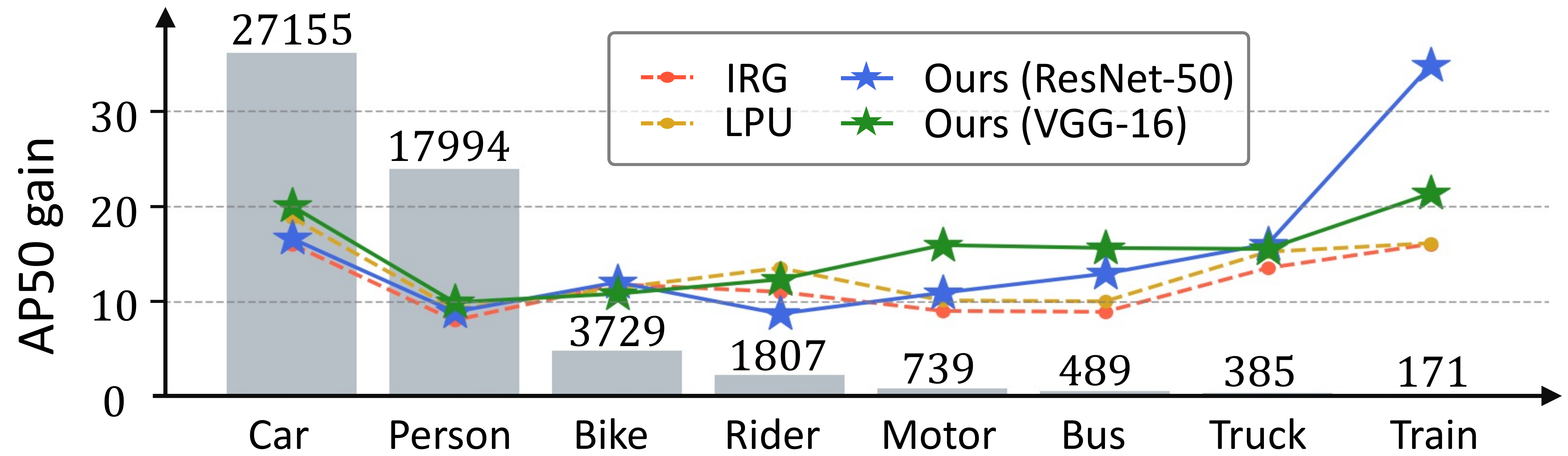}
    \caption{
    Number of instances and AP50 gain for each category.
    }
    \label{fig:each_category}
\end{figure}
On Fig.~\ref{fig:each_category}, we report the number of instances and the AP50 gain achieved by our method for each class compared to the source model on Cityscapes to Foggy Cityscapes, using ResNet-50~\cite{resnet} and VGG-16~\cite{vgg} backbones.
Since motor, bus, truck, train have lower number of instances compared to other classes, we refer to them as minor classes.
Our method significantly increases performance on minor classes and shows comparable gains to other methods in major classes.
\subsection{Analysis on Low-confidence Pseudo Label (LPL) mining}

\subsubsection{Class Alignment in LPL Mining Process}
After extracting High-confidence Pseudo Labels (HPL), our LPL mining proceeds three thresholding operations: IoU thresholding with HPL with $\delta_{IoU}$, background thresholding using $\delta_{bg}$, and confidence thresholding through $\delta_{lc}$ for choosing LPL.
\begin{figure}[t]
    \centering
    \includegraphics[width=\textwidth]{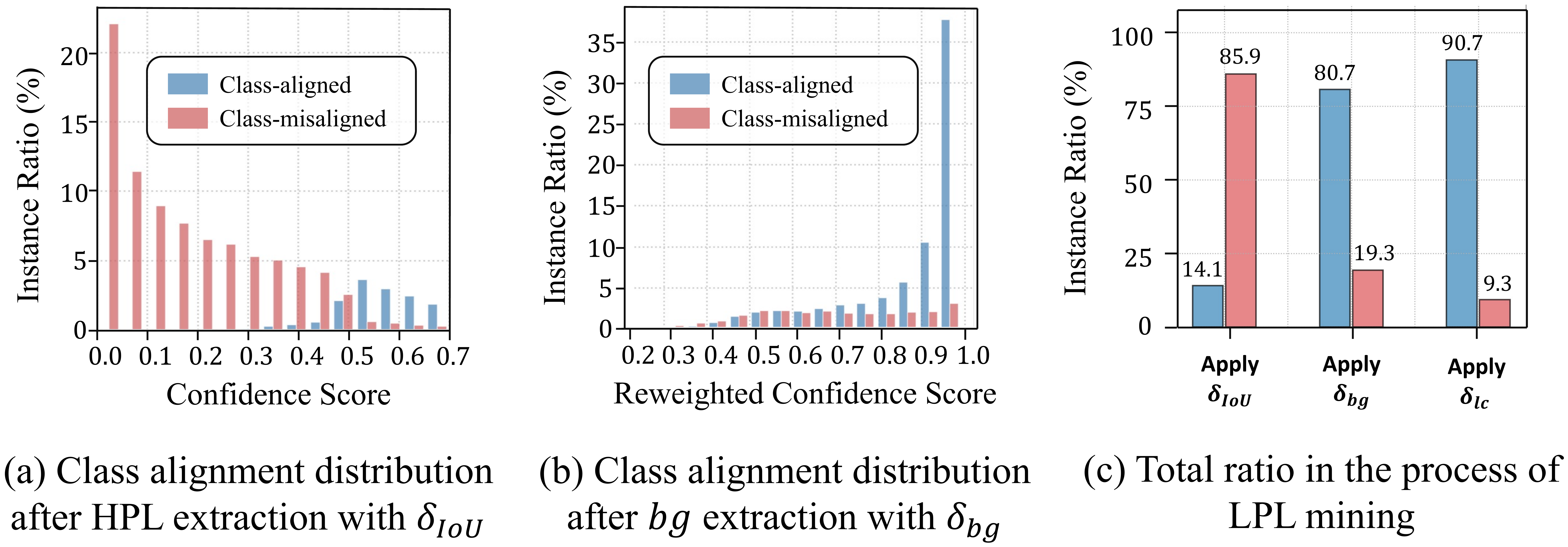}
    \caption{
    Analysis on the impact of reweighted confidence scores and the application of thresholding in the LPL mining process on class alignment.
    }
    \label{fig:pl_effect}
\end{figure}
As depicted in Fig.~\ref{fig:pl_effect} (a) and Fig.~\ref{fig:pl_effect} (c), only a few instances ($14.1\%$) after applying IoU threshold value $\delta_{IoU}$ are class-aligned. 
To address the class-misalignment issue in low-confidence proposals, we eliminate noisy bounding boxes by applying a background probability threshold and utilize a reweighted confidence score that excludes the background score. This approach markedly increases the proportion of class-aligned instances from 14.1\% to 80.7\%. Furthermore, by filtering out boxes using a confidence threshold in LPL mining, we achieve an additional enhancement in the ratio of correctly class-aligned instances, elevating it from 80.7\% to 90.7\%.

 \begin{figure}[htb!]
    \centering
    \includegraphics[width=\textwidth]{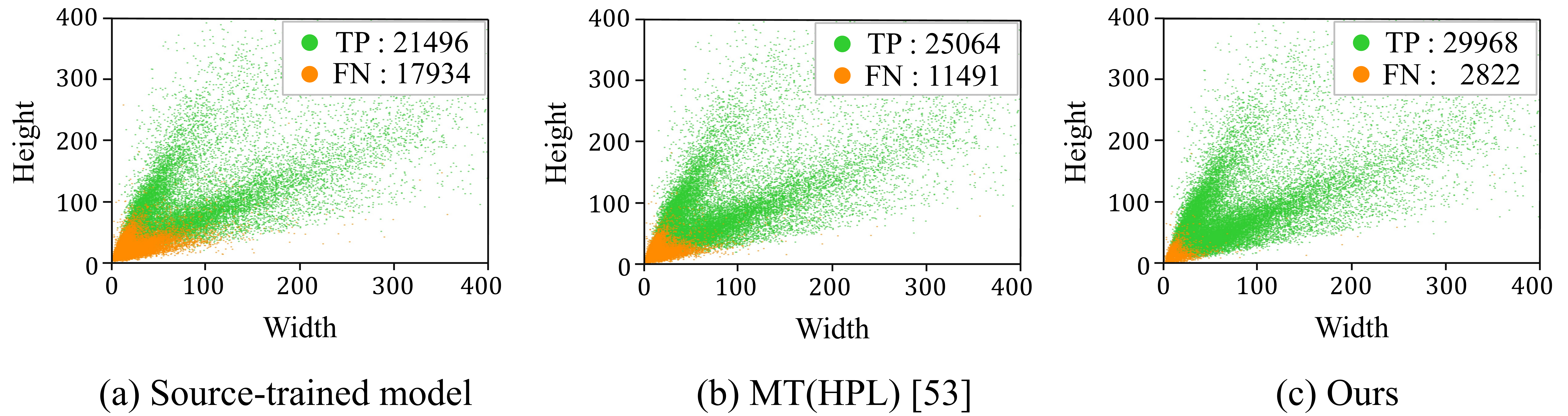}
    \caption{Comparison of True Positives, False Negatives on (a) Source-trained model, (b) MT(HPL)~\cite{mean_teacher}, (c) Ours. Height, Width represent the height and width of each instance. For better visualization, instances with their width, height less than 400 are displayed, as only a few instances over this size exist.}
    \label{fig:tpfn}
\end{figure}

\subsubsection{False Negatives vs. Instance Scale}
In Fig.~\ref{fig:tpfn}, we report the comparisons of the number, size of true positive instances and false nagative instances with respect to the Source-trained model, MT(HPL)\cite{mean_teacher}, and our LPLD model. 
Our method has 29968 true positives and 2822 false negatives, whereas MT has 25064 true positives and 11491 false negatives.
Although MT had fewer true positives, and more false negatives than the source-trained model, our method exhibited even more true positives and significantly less false negatives than the MT method.
This reduction in false negatives can be seen in the bottom left corner of Fig.~\ref{fig:tpfn}.
Note that most of the false negatives are small-sized instances.
Our method captures hard positive objects much better, such as small scale instances.

\begin{figure}[h]
    \begin{minipage}[h]{.325\linewidth}
    \centering
    \includegraphics[width=\linewidth]{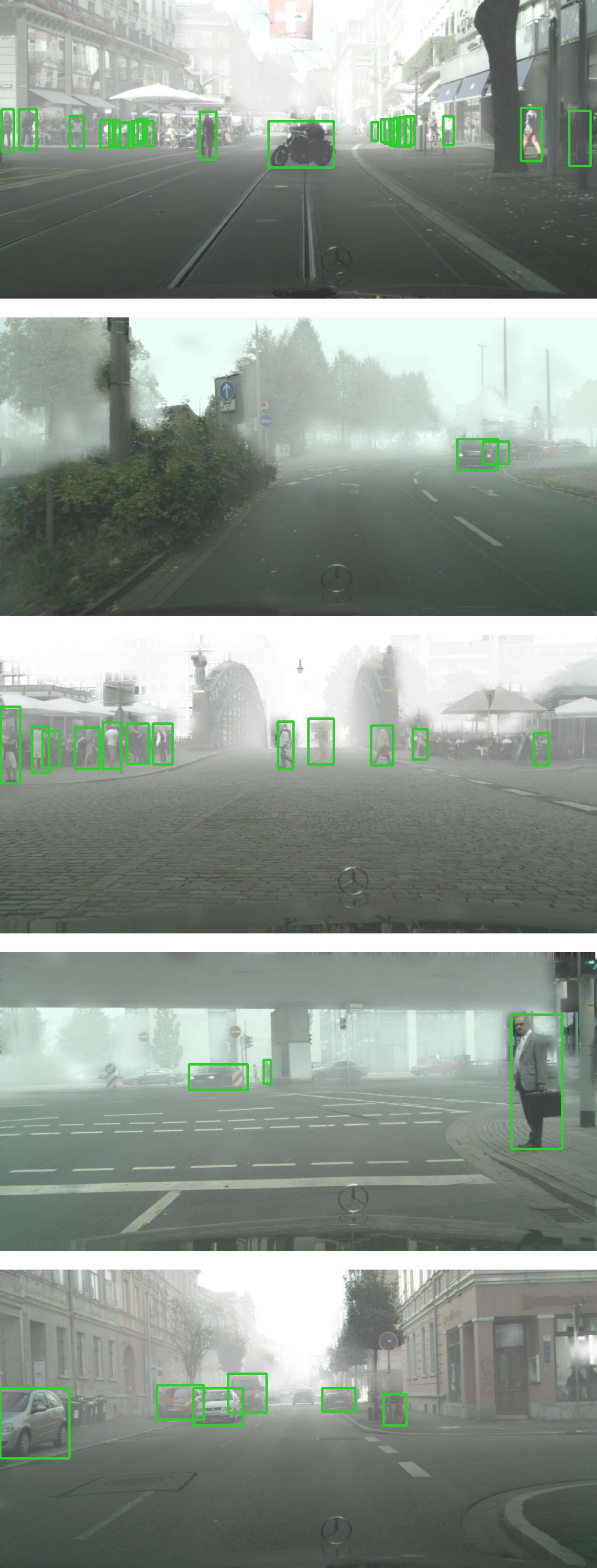}
    \caption*{\scriptsize{(a) HPL}}
  \end{minipage}
  \begin{minipage}[h]{.325\linewidth}
    \centering
    \includegraphics[width=\linewidth]{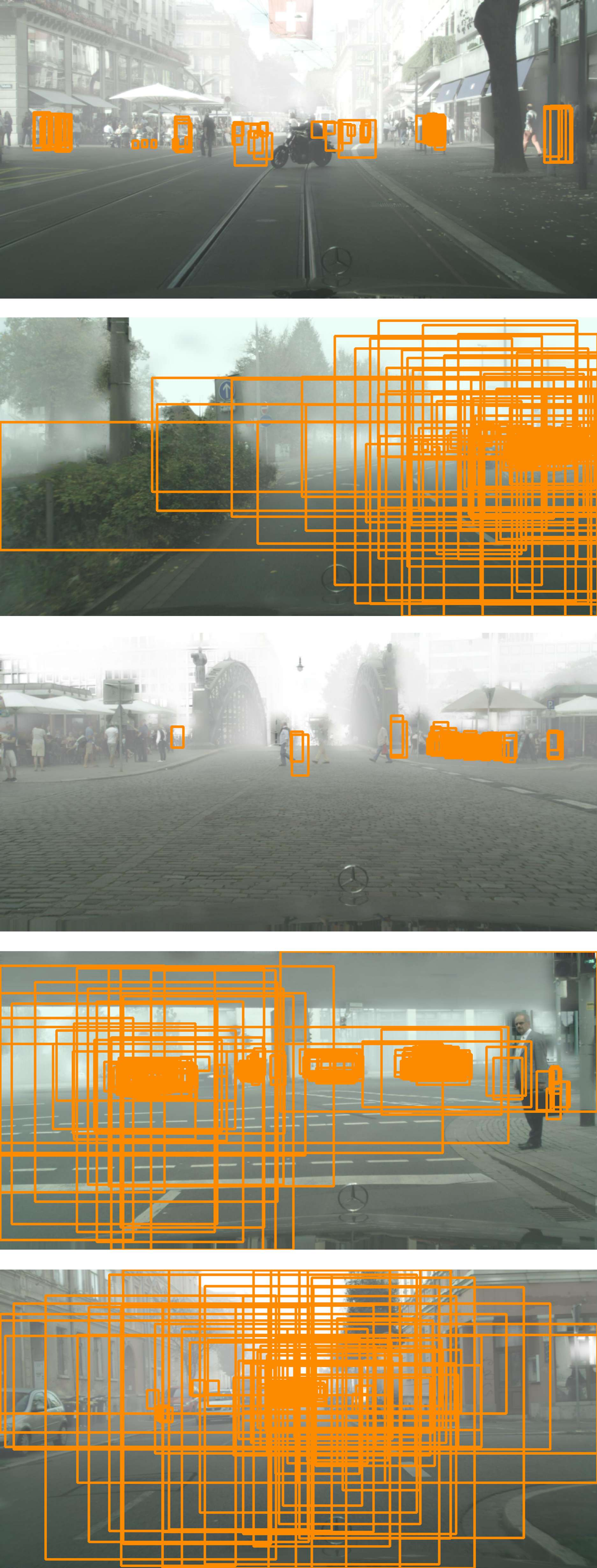}
    \caption*{\scriptsize{(b) LPL after applying $\delta_{IoU}$}}
  \end{minipage}
  \begin{minipage}[h]{.325\linewidth}
    \centering
    \includegraphics[width=\linewidth]{figure/visualization_/pl_visualization3_v2.pdf}
    \caption*{\scriptsize{(c) LPL after applying $\delta_{bg}$, $\delta_{lc}$}}
  \end{minipage}
  \caption{Visualization of High-confidence Pseudo Labels (HPL) and Low-confidence Pseudo Labels (LPL) in our LPL mining process in Cityscapes\cite{cityscapes} $\rightarrow$ Foggy Cityscapes\cite{foggy} scenario.}
  \label{fig:qual_pl}
\end{figure}
\subsubsection{Visualization of HPL and LPL}
Visualization of HPL on Fig.~\ref{fig:qual_pl} (a) and LPL on Fig.~\ref{fig:qual_pl} (c) shows that HPL and LPL are capturing different foregrounds in the image, with HPL capturing objects that are easier for the model to detect compared to the objects that LPL captures. 
Comparison of Fig.~\ref{fig:qual_pl} (b) and Fig.~\ref{fig:qual_pl} (c) shows that applying our LPL mining process, especially $\delta_{bg}$ and $\delta_{lc}$ thresholds, is crucial for eliminating numerous non-foreground boxes following the thresholding using $\delta_{IoU}$.

\section{Additional Qualitative Results}\label{sec:supplequal}
\begin{figure}[htb!]
    \begin{minipage}[h]{.245\linewidth}
    \centering
    \includegraphics[width=\linewidth]{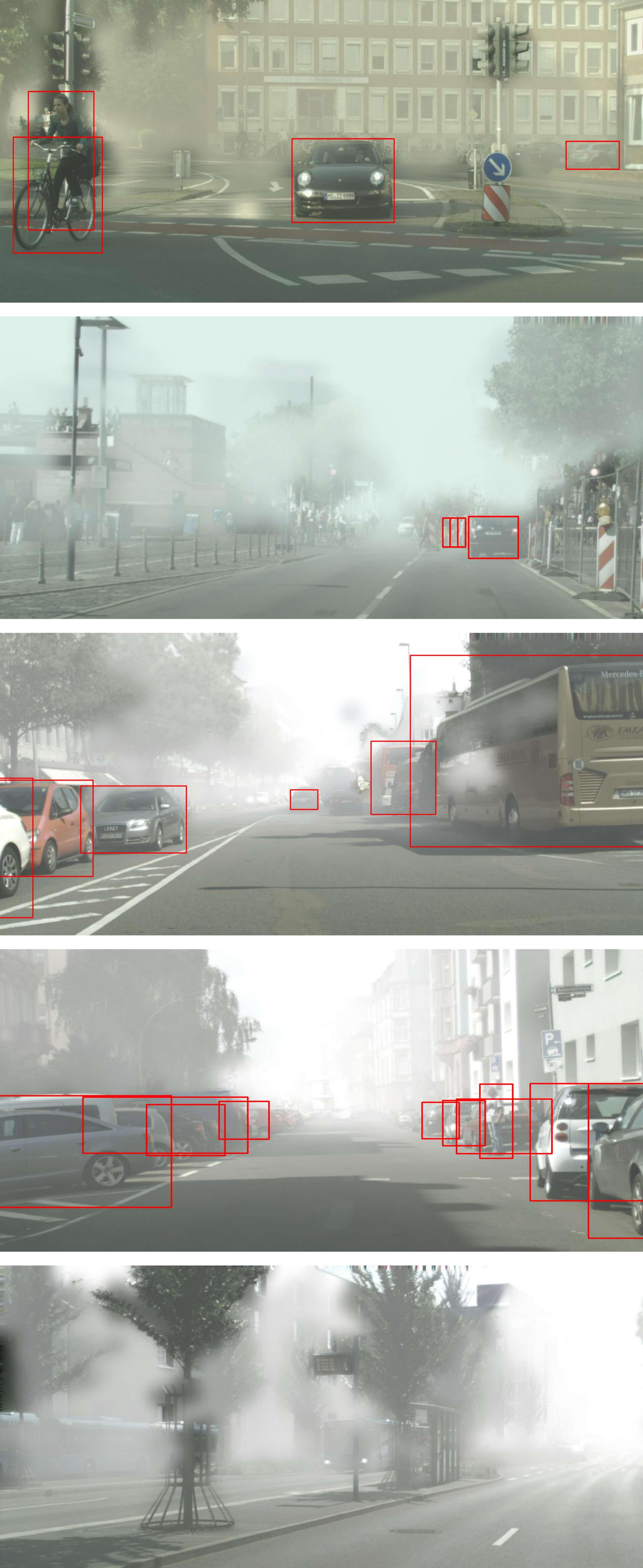}
    \caption*{\scriptsize{(a) Source}}
  \end{minipage}
  \begin{minipage}[h]{.245\linewidth}
    \centering
    \includegraphics[width=\linewidth]{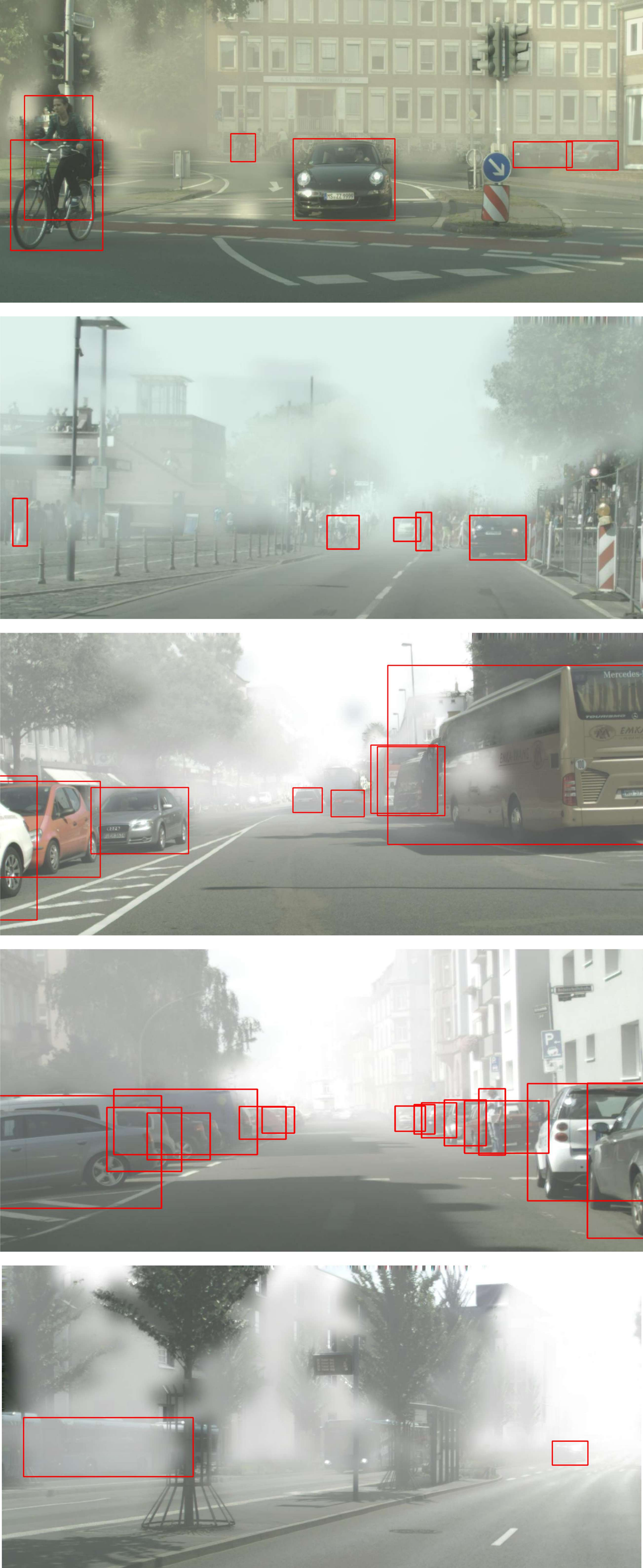}
    \caption*{\scriptsize{(b) MT(HPL)~\cite{mean_teacher}}}
  \end{minipage}
  \begin{minipage}[h]{.245\linewidth}
    \centering
    \includegraphics[width=\linewidth]{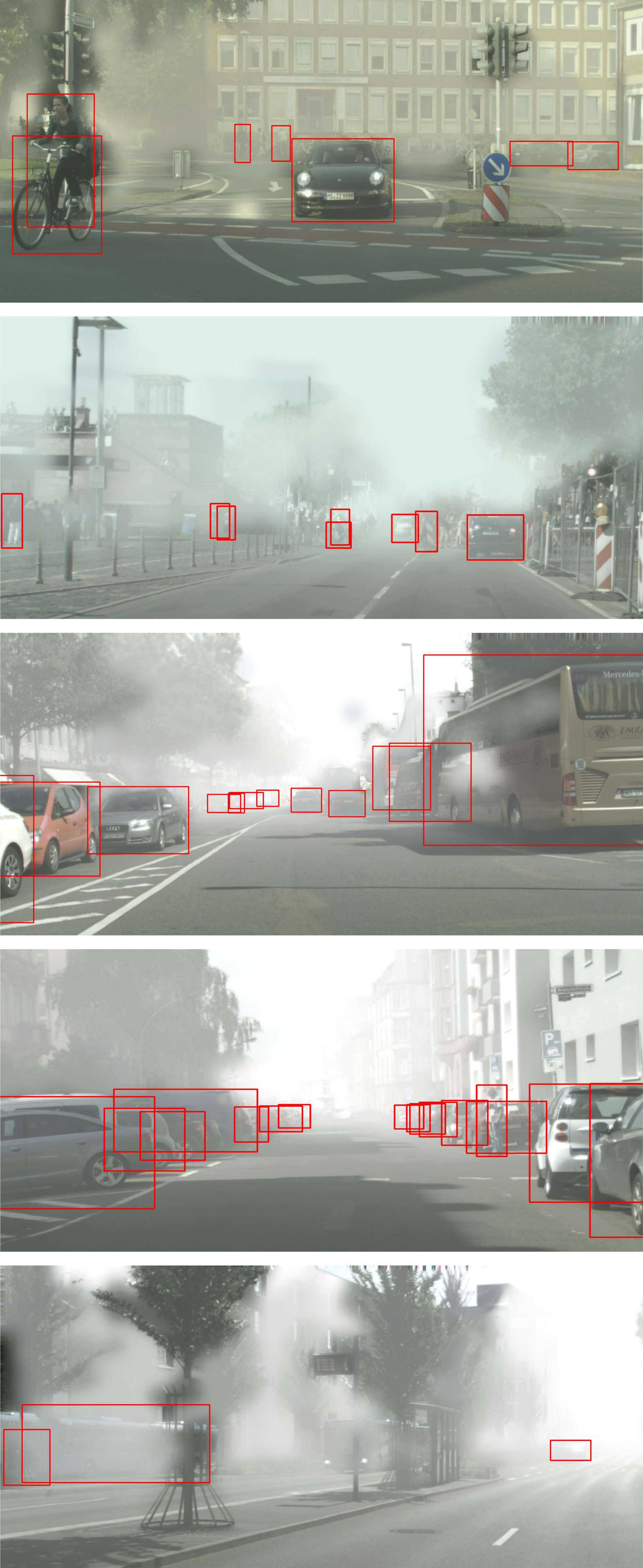}
    \caption*{\scriptsize{(c) IRG~\cite{irg_sfda}}}
  \end{minipage}
  \begin{minipage}[h]{.245\linewidth}
    \centering
    \includegraphics[width=\linewidth]{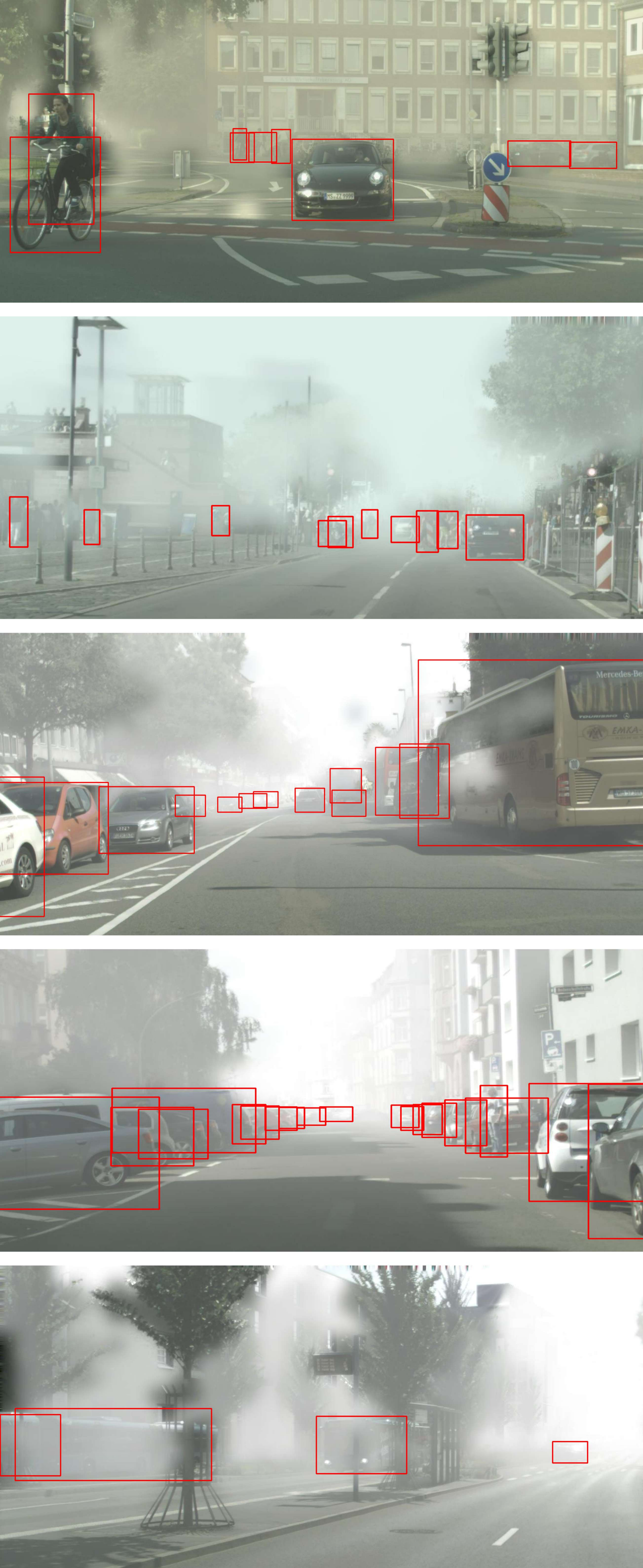}
    \caption*{\scriptsize{(d) Ours}}
  \end{minipage}
  \caption{Additional qualitative results for Cityscapes\cite{cityscapes} $\rightarrow$ Foggy Cityscapes\cite{foggy}. Bounding boxes in \textbf{\color{red}red} refer to the prediction. \textit{Zoom in for best view.}}
  \label{fig:qual_foggy2}
\end{figure}

\begin{figure}[t]
    \begin{minipage}[h]{.245\linewidth}
    \centering
    \includegraphics[width=\linewidth]{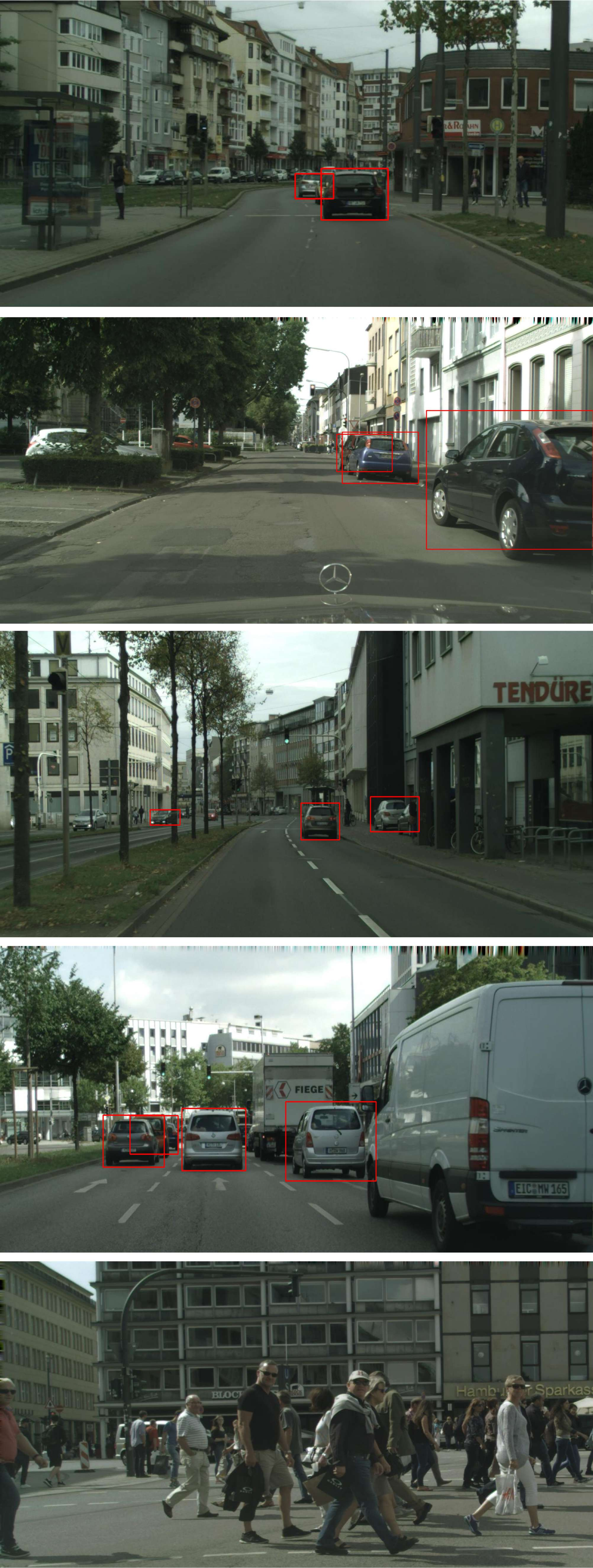}
    \caption*{\scriptsize{(a) Source}}
  \end{minipage}
  \begin{minipage}[h]{.245\linewidth}
    \centering
    \includegraphics[width=\linewidth]{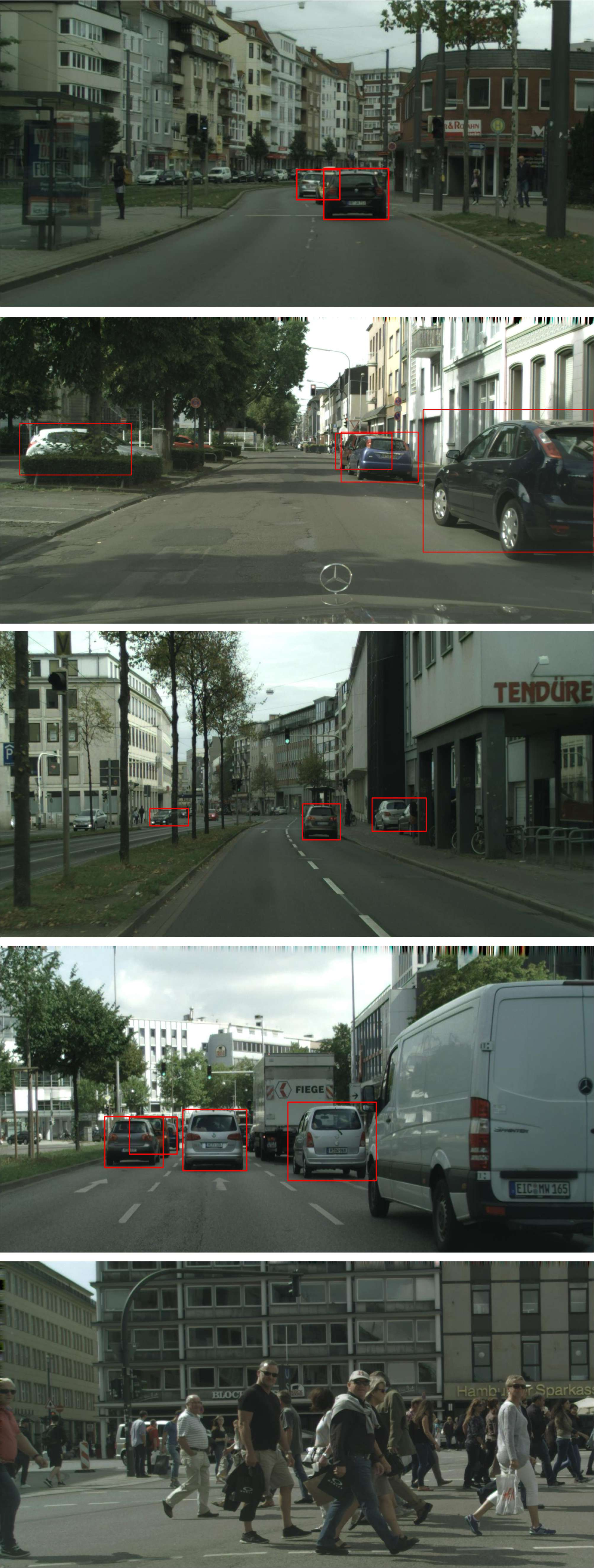}
    \caption*{\scriptsize{(b) MT(HPL)~\cite{mean_teacher}}}
  \end{minipage}
  \begin{minipage}[h]{.245\linewidth}
    \centering
    \includegraphics[width=\linewidth]{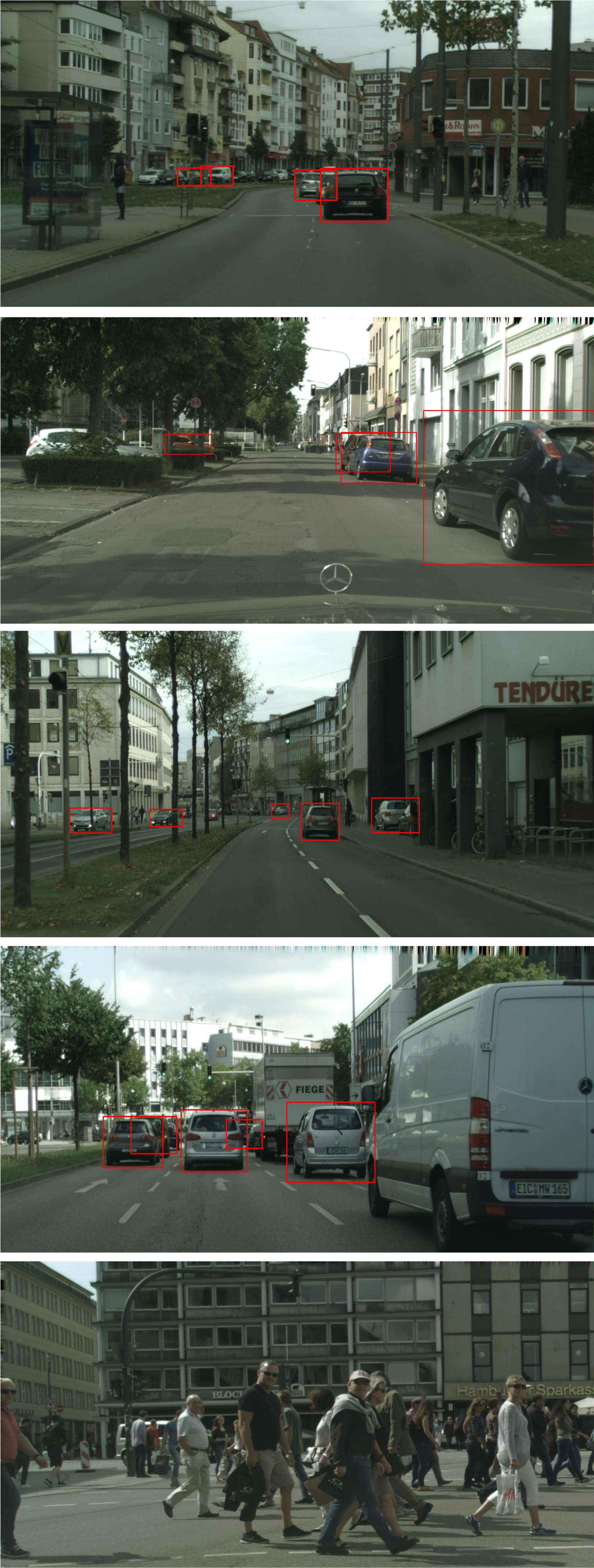}
    \caption*{\scriptsize{(c) IRG~\cite{irg_sfda}}}
  \end{minipage}
  \begin{minipage}[h]{.245\linewidth}
    \centering
    \includegraphics[width=\linewidth]{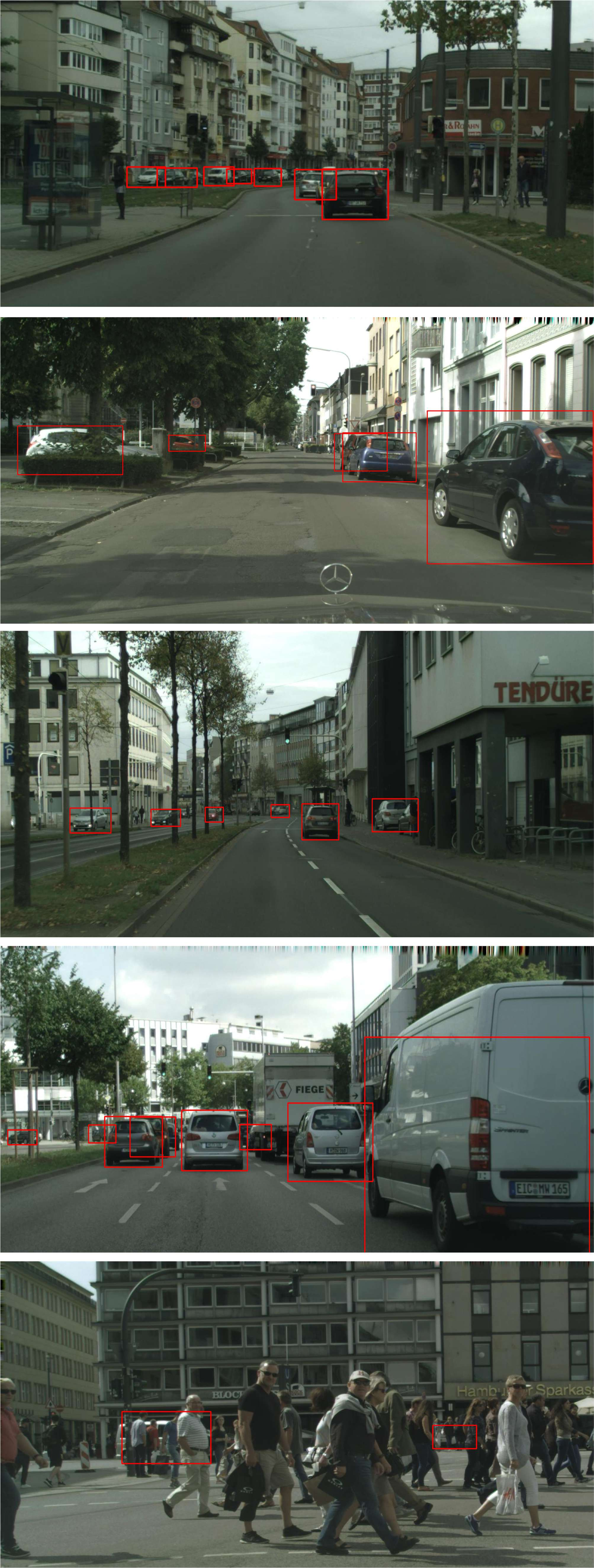}
    \caption*{\scriptsize{(d) Ours}}
  \end{minipage}
  \caption{Qualitative results for Kitti\cite{kitti} $\rightarrow$ Cityscapes\cite{cityscapes} Bounding boxes in \textbf{\color{red}red} refer to the prediction. \textit{Zoom in for best view.}}
  \label{fig:qual_kitti}
\end{figure}

\begin{figure}[h]
    \begin{minipage}[h]{.245\linewidth}
    \centering
    \includegraphics[width=\linewidth]{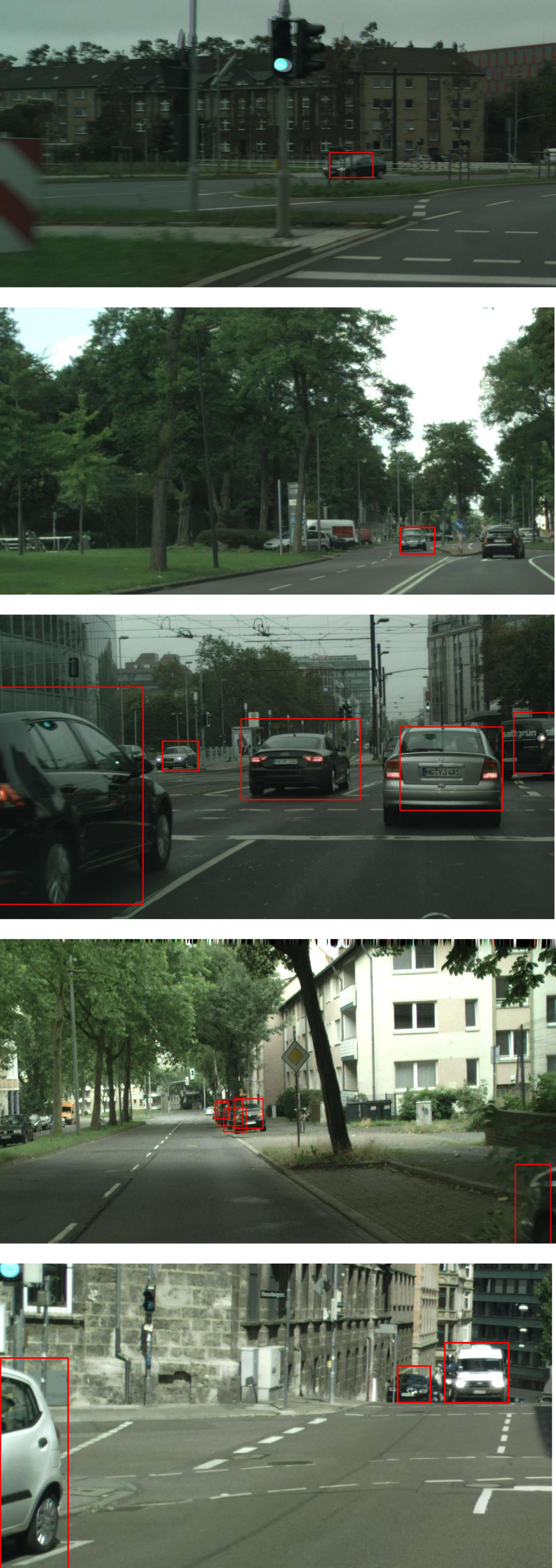}
    \caption*{\scriptsize{(a) Source}}
  \end{minipage}
  \begin{minipage}[h]{.245\linewidth}
    \centering
    \includegraphics[width=\linewidth]{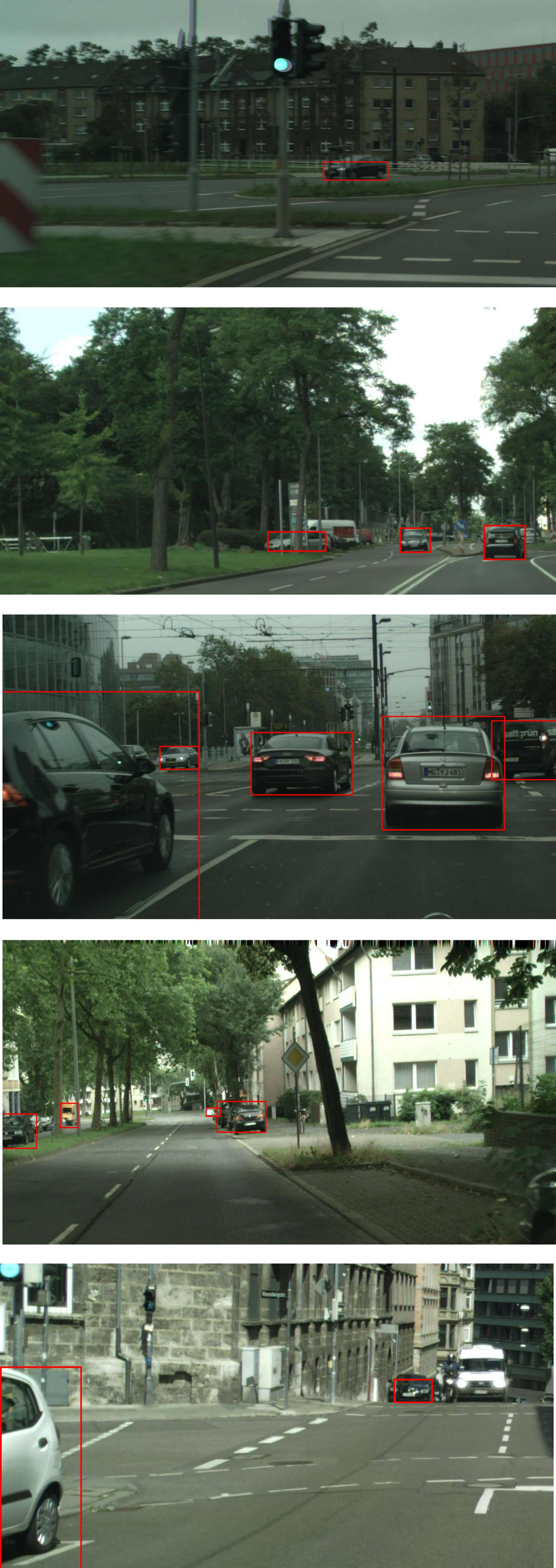}
    \caption*{\scriptsize{(b) MT(HPL)~\cite{mean_teacher}}}
  \end{minipage}
  \begin{minipage}[h]{.245\linewidth}
    \centering
    \includegraphics[width=\linewidth]{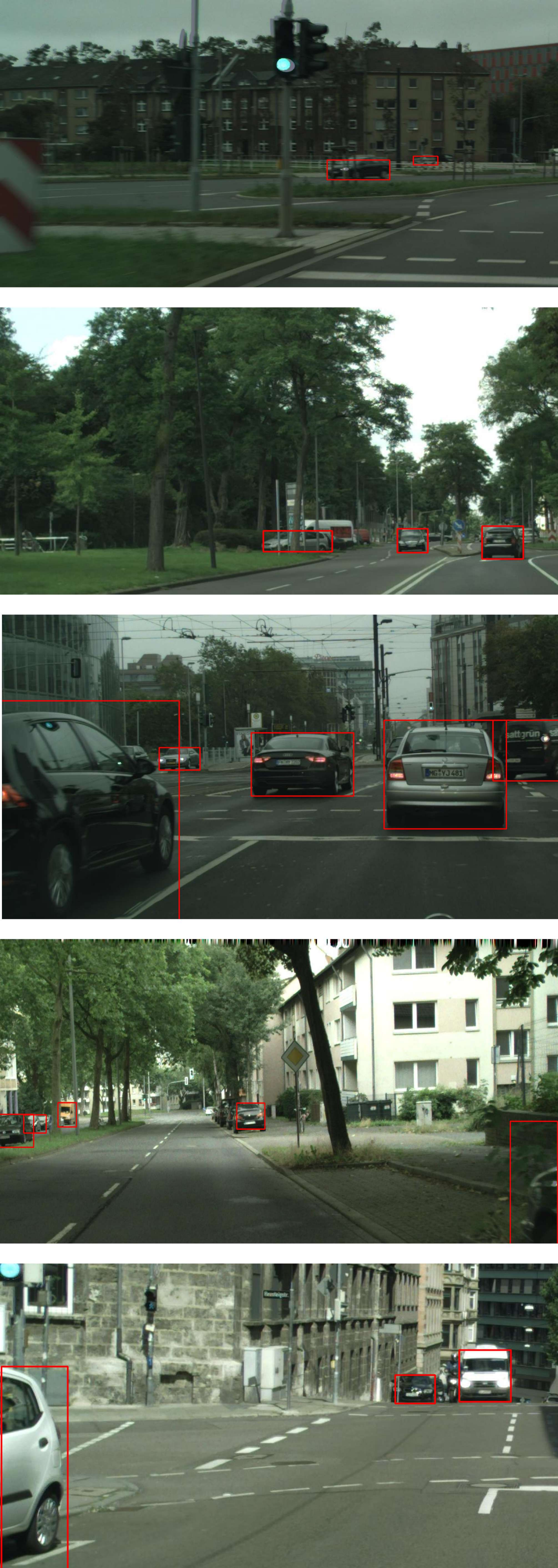}
    \caption*{\scriptsize{(c) IRG~\cite{irg_sfda}}}
  \end{minipage}
  \begin{minipage}[h]{.245\linewidth}
    \centering
    \includegraphics[width=\linewidth]{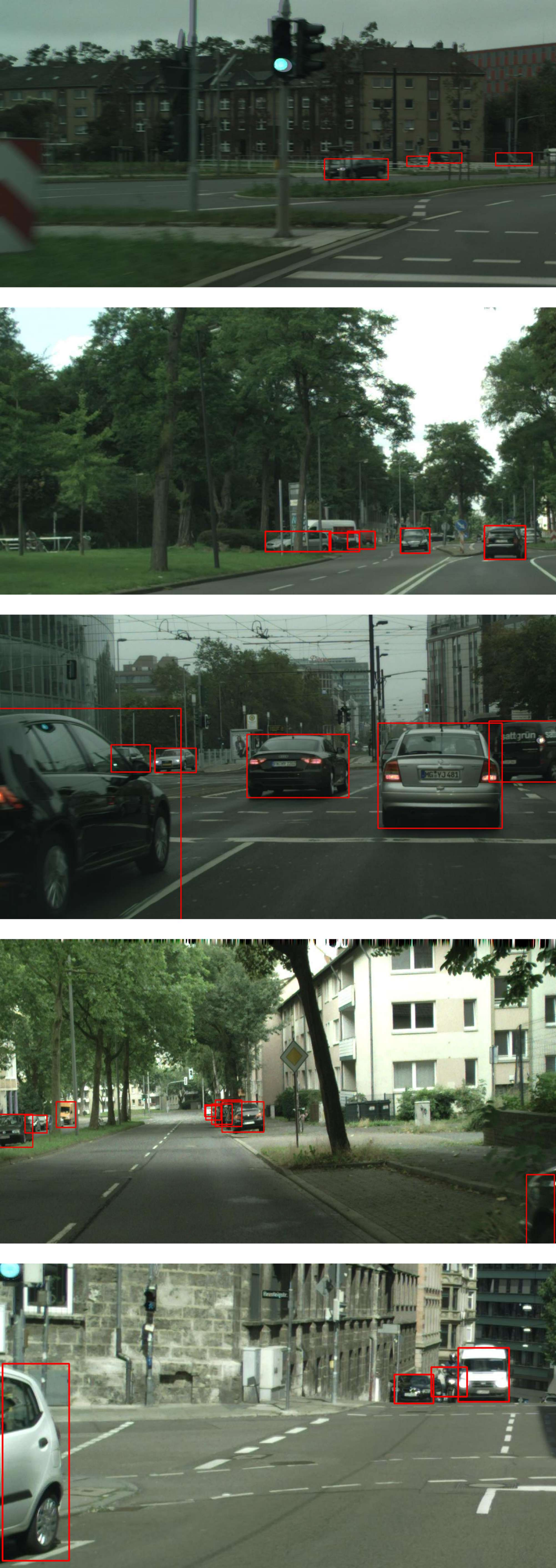}
    \caption*{\scriptsize{(d) Ours}}
  \end{minipage}
  \caption{Qualitative results for Sim10k\cite{sim10k} $\rightarrow$ Cityscapes\cite{cityscapes} Bounding boxes in \textbf{\color{red}red} refer to the prediction. \textit{Zoom in for best view.}}
  \label{fig:qual_sim}
\end{figure}

\begin{figure}[htb!]
    \begin{minipage}[h]{.245\linewidth}
    \centering
    \includegraphics[width=\linewidth]{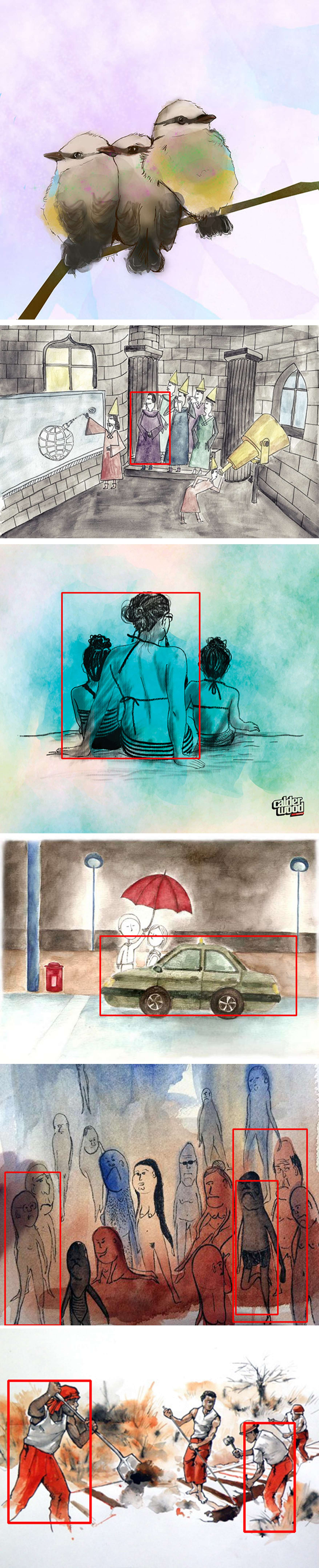}
    \caption*{\scriptsize{(a) Source}}
  \end{minipage}
  \begin{minipage}[h]{.245\linewidth}
    \centering
    \includegraphics[width=\linewidth]{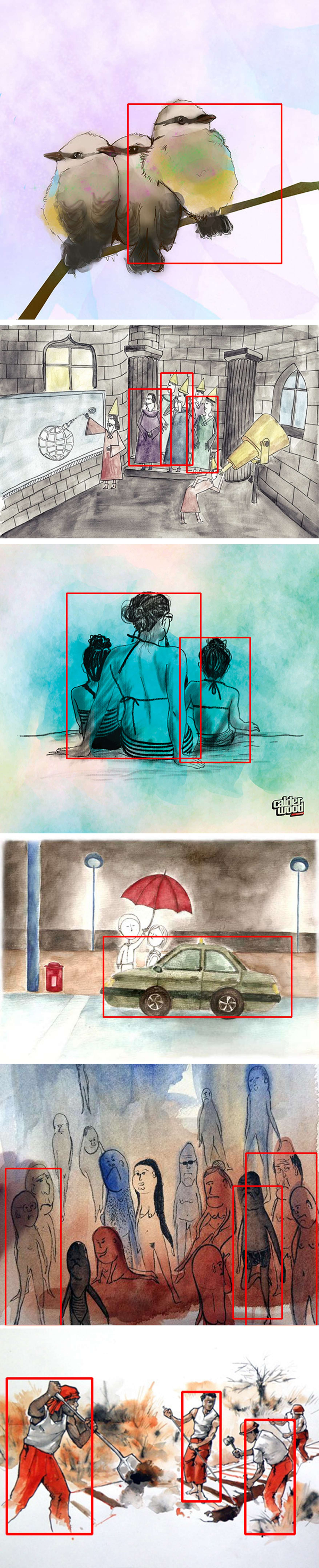}
    \caption*{\scriptsize{(b) MT(HPL)~\cite{mean_teacher}}}
  \end{minipage}
  \begin{minipage}[h]{.245\linewidth}
    \centering
    \includegraphics[width=\linewidth]{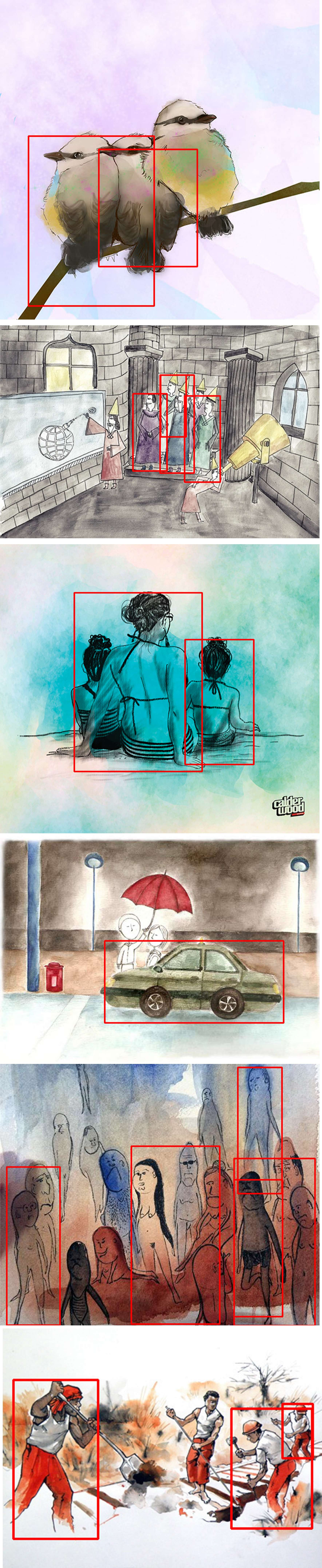}
    \caption*{\scriptsize{(c) IRG~\cite{irg_sfda}}}
  \end{minipage}
  \begin{minipage}[h]{.245\linewidth}
    \centering
    \includegraphics[width=\linewidth]{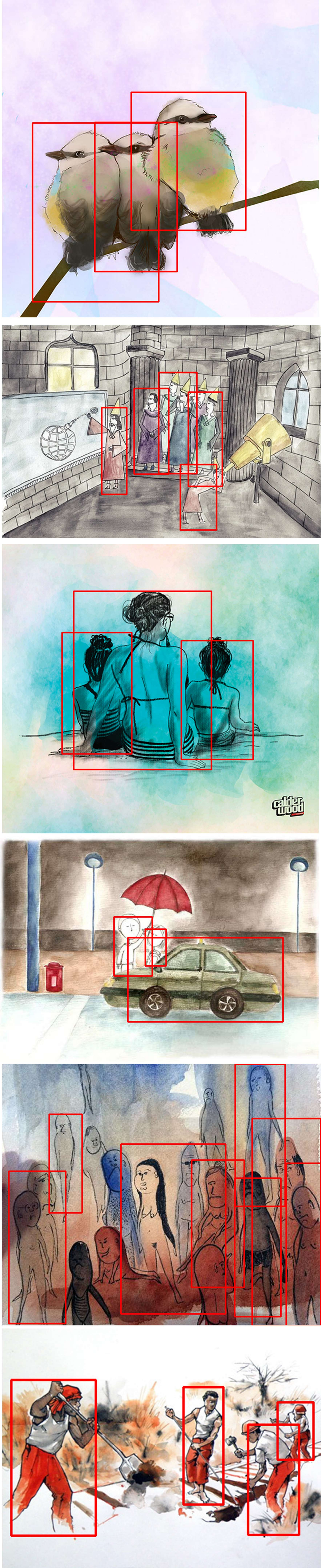}
    \caption*{\scriptsize{(d) Ours}}
  \end{minipage}
  \caption{Qualitative results for Pascal-VOC\cite{pascal} $\rightarrow$ Watercolor\cite{clip_water} Bounding boxes in \textbf{\color{red}red} refer to the prediction.}
  \label{fig:qual_water}
\end{figure}

\begin{figure}[htb!]
    \begin{minipage}[h]{.245\linewidth}
    \centering
    \includegraphics[width=\linewidth]{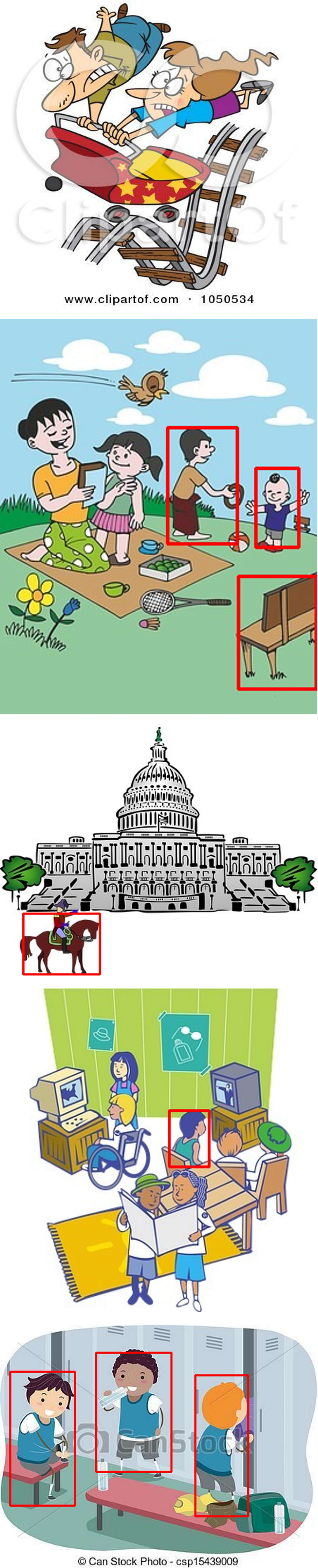}
    \caption*{\scriptsize{(a) Source}}
  \end{minipage}
  \begin{minipage}[h]{.245\linewidth}
    \centering
    \includegraphics[width=\linewidth]{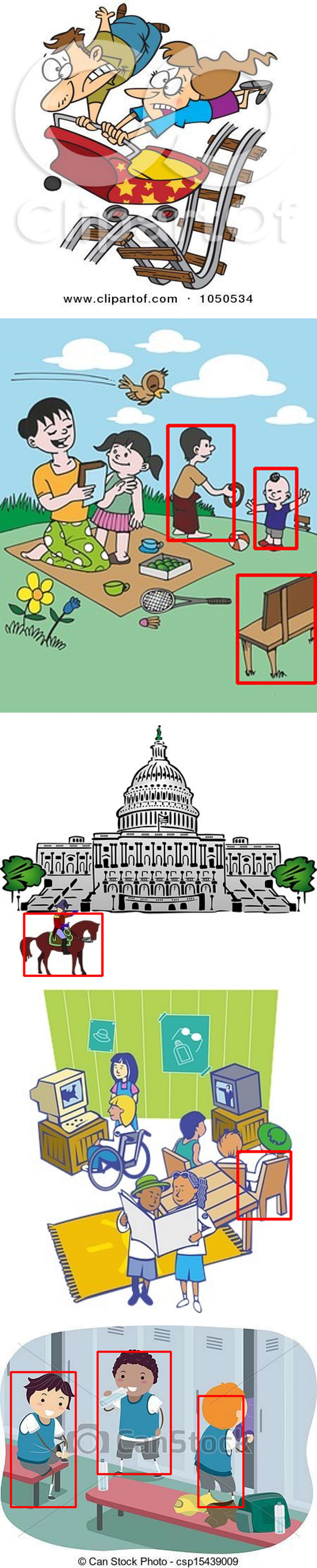}
    \caption*{\scriptsize{(b) MT(HPL)~\cite{mean_teacher}}}
  \end{minipage}
  \begin{minipage}[h]{.245\linewidth}
    \centering
    \includegraphics[width=\linewidth]{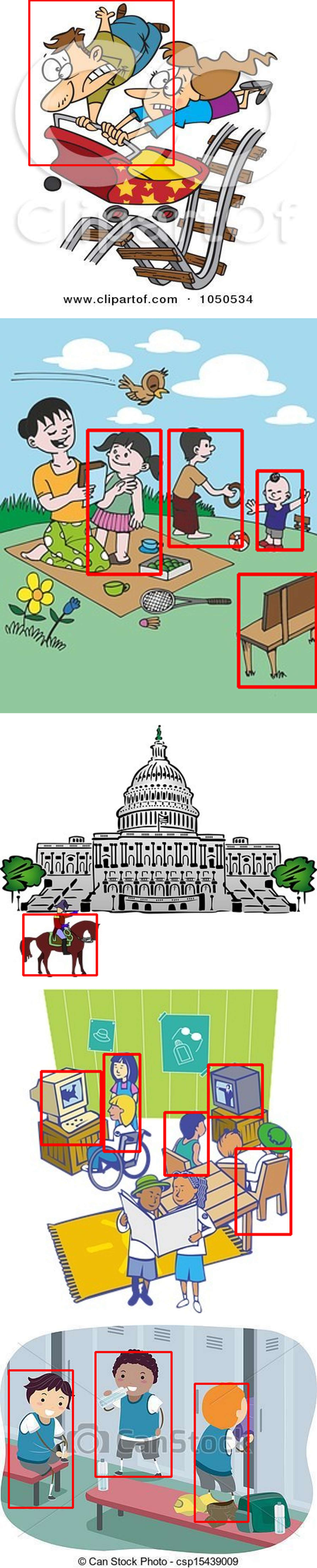}
    \caption*{\scriptsize{(c) IRG~\cite{irg_sfda}}}
  \end{minipage}
  \begin{minipage}[h]{.245\linewidth}
    \centering
    \includegraphics[width=\linewidth]{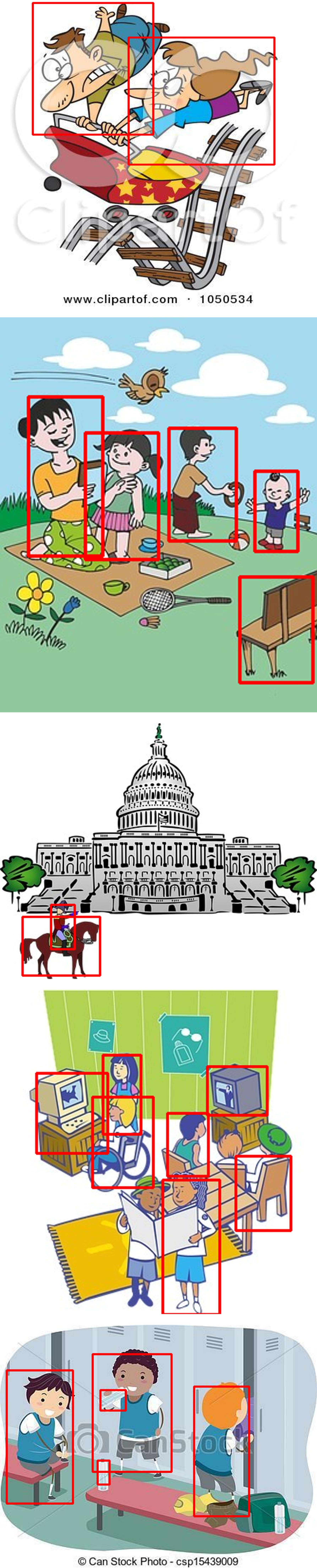}
    \caption*{\scriptsize{(d) Ours}}
  \end{minipage}
  \caption{Qualitative comparison for Pascal-VOC\cite{pascal} $\rightarrow$ Clipart\cite{clip_water} Bounding boxes in \textbf{\color{red}red} refer to the prediction.}
  \label{fig:qual_clip}
\end{figure}

For a more comprehensive understanding, we provide additional qualitative results on diverse domain shift scenarios, including Cityscapes\cite{cityscapes} to Foggy Cityscapes\cite{foggy}, Kitti\cite{kitti} to Cityscapes, Sim10k\cite{sim10k} to Cityscapes, Pascal-VOC\cite{pascal} to Watercolor\cite{clip_water}, and Pascal-VOC to Clipart\cite{clip_water}, as shown in Fig. \labelcref{fig:qual_foggy2,fig:qual_clip,fig:qual_water,fig:qual_sim,fig:qual_kitti}.
On all domain shift scenarios, we compare our model's result with source pre-trained model, Mean-Teacher\cite{mean_teacher}, and IRG\cite{irg_sfda}.

In Fig.~\ref{fig:qual_foggy2}, we show the detection result on Cityscapes to Foggy Cityscapes, which is a weather change scenario.
In a foggy environment, small objects and objects occluded by either fog or other entities tend to exhibit low confidence, which may result in their exclusion from the training.
Owing to our LPLD's progressive exploration of these false negatives, our model successfully identifies these objects in contrast to alternative methods.
We also show the detection result from Kitti to Cityscapes, in Fig.~\ref{fig:qual_kitti}.
Similar to the previous scenario, our model can find occluded objects and small objects that other methods struggle with.

Fig.~\ref{fig:qual_sim} shows the detection result on Sim10k to Cityscapes, which is a simulation to real domain shift scenario.
Our method is better than other methods for detecting small or occluded objects.
We conjecture that our method can effectively handle the severe texture variations caused by sim-to-real domain shift, particularly for small or occluded objects.

Fig.~\ref{fig:qual_water}, ~\ref{fig:qual_clip} shows the detection result from Pascal-VOC to Watercolor, Clipart.
In both domains, objects are depicted in a totally different manner from their real-world counterparts (\eg Pascal-VOC), despite being categorized identically. 
Moreover, there is substantial variability in the appearance of objects belonging to the same category across different images within the same dataset.
This variability results in numerous instances receiving low confidence, regardless of their size.
By leveraging LPL, our model effectively incorporates objects with significant variance into training.
Overall results demonstrate the effectiveness of our method for the detection capability on various domain shifts.

\end{document}